\documentclass[
12pt,                  
a4paper,               
titlepage,             
headsepline,
appendixprefix,        
headings=normal,       
parskip=half*,         
captions=tableheading, 
chapterprefix,         
oneside,
BCOR=15mm,             
bibliography=totoc,
numbers=noenddot,
version
]{scrbook}

\usepackage{ucs}
\usepackage[utf8]{inputenc}
\usepackage[pdftex]{graphicx}
\usepackage{color}
\usepackage[ngerman, english]{babel}
\usepackage[automark]{scrpage2}  

\usepackage[usenames,dvipsnames]{xcolor}
\usepackage{tabularx}
\newcolumntype{R}[1]{>{\raggedleft\arraybackslash}p{#1}} 
\usepackage[titletoc]{appendix}
\usepackage{url}
\usepackage[colorlinks=false, pdfborder={0 0 0}]{hyperref}

\usepackage{hyperref}
\usepackage{caption}
\newlength\Colsep
\usepackage{dsfont}
\usepackage{amsmath}
\usepackage{amstext}

\usepackage{verbatim}
\usepackage{algorithm}
\usepackage{algorithmic}
\floatname{algorithm}{\footnotesize Algorithmus}
\usepackage{booktabs}

\usepackage{graphicx}
\usepackage{fancyvrb} 
\usepackage{siunitx}
\usepackage{listings}
\usepackage{tabularx}
\usepackage{colortbl}
\usepackage{hhline}
\usepackage{changepage}
\usepackage{mathtools}

\usepackage{subcaption}

\usepackage{bibgerm}
\usepackage{hyperref}
\DeclareOldFontCommand{\sc}{\normalfont\scshape}{\@nomath\sc}

\setkomafont{captionlabel}{\bfseries\footnotesize}
\setkomafont{caption}{\footnotesize}

\usepackage{calc}
\usepackage{geometry}
\usepackage[%
       final,%
       activate,%
       verbose=true]{microtype}
\usepackage{%
       mparhack%
}

\definecolor{Ocean}{cmyk}{1,0,0.2,0.78}
\definecolor{Grey}{cmyk}{0,0,0,0.6}
\definecolor{ULGray}{cmyk}{0,0,0.03,0.1}

\clearscrheadfoot
\rehead{\headmark} 
\lehead{\pagemark} 
\lohead{\headmark} 
\rohead{\pagemark}

\usepackage{booktabs}
\usepackage{tabularx}
\usepackage{colortbl}
\usepackage{hhline}

\usepackage[pdftex]{graphicx}
\usepackage{caption}
\usepackage{threeparttable}

\usepackage{multirow}

\begin{document}

\thispagestyle{headings}
\pagenumbering{roman}
\begin{titlepage}
\vspace*{\stretch{1}}

\addtolength{\topmargin}{-1.2cm} 
\addtolength{\textwidth}{2.35cm} 

\vspace*{-2.7cm}
\includegraphics[width=0.65\textwidth]{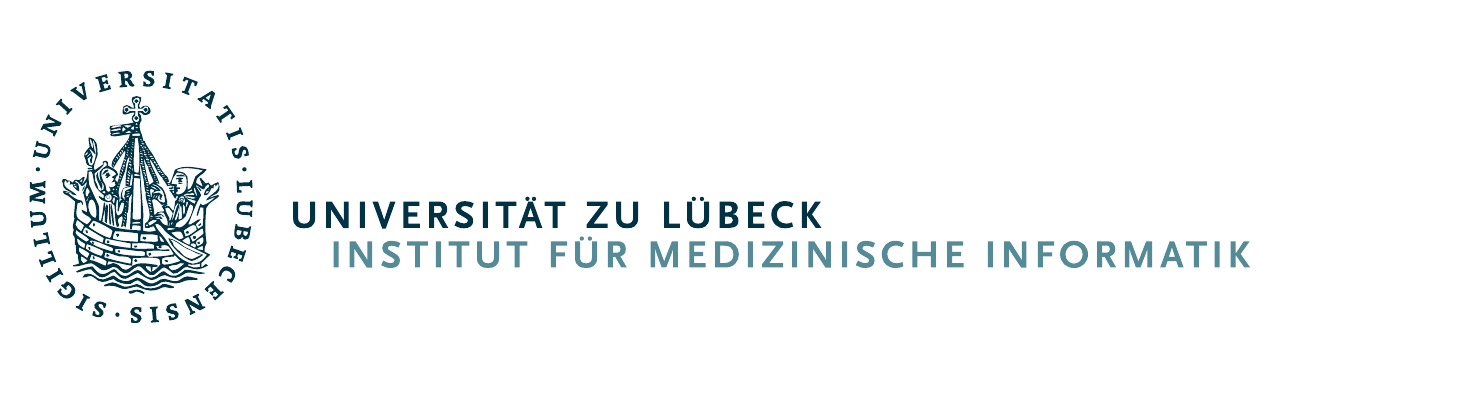} 

\vspace*{0.4cm}
\begin{center}


\enlargethispage{5cm}
Aus dem Institut für Medizinische Informatik 
der Universität zu Lübeck\\
Direktor: Prof. Dr. rer. nat. habil. Heinz Handels\\[1.8cm]

\begin{LARGE}
\textcolor{Ocean}{{\textbf{CNNs, LSTMs, and Attention Networks for Pathology Detection in Medical Data}}}\\
\end{LARGE}
\vspace*{1.5cm} 

{{\textbf{CNNs, LSTMs und Attention Netzwerke für die Detektion von Pathologien in medizinischen Daten} }}\\

\vspace*{2.5cm} 

Masterarbeit \\
im Rahmen des Studienganges Medizinische Informatik\\ der Universität zu Lübeck\\[1.0cm]

vorgelegt von\\[0.1cm]
\textbf{Nora Vogt}\\[1.0cm]

ausgegeben und betreut von\\[0.1cm] 

{\textbf{Prof. Dr. Mattias P. Heinrich}}\\[0.3cm]

mit Unterstützung von\\[0.1cm] 

{\textbf{Dr. Julien Oster}\textsuperscript{1}}\\[0.5cm]

\textsuperscript{1} IADI U1254, Chargé de Recherche (Senior Research Fellow) INSERM\\

\vspace*{2cm}

Lübeck, den 24. Juli 2018

\end{center}
  

\vspace*{\stretch{5}}
\newpage
\thispagestyle{empty}

\addtolength{\topmargin}{1.2cm} 
\addtolength{\textwidth}{-2.35cm} 

\end{titlepage}


\cleardoublepage
\chapter*{Abstract} 

For the weakly supervised task of electrocardiogram (ECG) rhythm classification, convolutional neural networks (CNNs) and long short-term memory (LSTM) networks are two increasingly popular classification models. This work investigates whether a combination of both architectures to so-called convolutional long short-term memory (ConvLSTM) networks can improve classification performances by explicitly capturing morphological as well as temporal features of raw ECG records. In addition, various attention mechanisms are studied to localize and visualize record sections of abnormal morphology and irregular rhythm. The resulting saliency maps are supposed to not only allow for a better network understanding but to also improve clinicians' acceptance of automatic diagnosis in order to avoid the technique being labeled as a black box. In further experiments, attention mechanisms are actively incorporated into the training process by learning a few additional attention gating parameters in a CNN model. An 8-fold cross validation is finally carried out on the PhysioNet Computing in Cardiology (CinC) challenge 2017 to compare the performances of standard CNN models, ConvLSTMs, and attention gated CNNs.

\tableofcontents



\cleardoublepage

\setcounter{page}{1}
\pagenumbering{arabic}

\chapter{Introduction}

As the detection of atrial fibrillation (AF) consists in the examination of long-term electrocardiograms (ECGs), the diagnosis of cardiac arrhythmias can be very time-consuming. Especially in cases of paroxysmal AF, episodes of AF might occur infrequently and therefore require heart activity records over several days. The aim of this work is the development of an automatic system for the detection of cardiac arrhythmias that can yield a fast and accurate diagnosis. Different visualization approaches are proposed to improve the interpretability and to experimentally validate classification outputs. It is suggested that the highlighting of salient ECG sections has the potential to not only lead to a better acceptance of several compared deep learning models but to also facilitate the subsequent inspection of clinicians in cases of high prediction uncertainty. 

The next section will give a brief overview of existing state-of-the-art ECG rhythm classifiers, which in many cases rely on (well designed) 
handcrafted features. Following the trend of deep learning, we try to evade the engineering part by automatically extracting a discriminative data representation with neural network architectures. Two popular architectures for the given task are convolutional neural networks (CNNs) \cite{lecun1998} and recurrent neural networks (RNNs) \cite{elman1990}/ long short-term memory networks (LSTMs) \cite{hochreiter1997}. While CNNs have proven to be particularly successful in image classification tasks (by learning translation invariant filters for pattern recognition), RNNs recently yielded promising achievements for the processing of temporal data (by learning short and long-range dependencies at different scales \cite{geras2015}). 

In this work, CNNs and LSTMs will be combined with the goal to first learn discriminative patterns and then discover temporal dependencies in a feature sequence. Besides studying a number of attention visualization approaches, an attention network is presented that is actively incorporating saliency information into the learning process.

\section{Motivation}

Atrial fibrillation is the most common cardiac arrhythmia with a population prevalence of 1-2 \% \cite{lip2016} 
and is known to be associated with a number of cardiovascular diseases (causing heart failure, ischemia, stroke, and even cardiac death \cite{odutayo2016}). The treatment of AF (drugs or ablation procedure) should be started as early as possible which implies that accurate diagnostic tools are strongly needed. To encourage the development of automatic diagnosis systems, the PhysioNet community \cite{goldberger2000} provides access to a large collection of physiological (and particularly cardiac) databases. For the PhysioNet Computing in Cardiology (CinC) challenge 2017 \cite{clifford2017} 8528 short single lead ECG recordings were made available for the purpose of AF classification. Even though the challenge is over by now, the hidden test set is kept private for further submissions. The CinC dataset (each record being annotated with one global rhythm label) will be used to evaluate the performance of our models. For a better validation of our attention visualization approaches, we furthermore considered the MIT-BIH database \cite{moody2001} (provided by the MIT and Boston's Beth Israel Hospital) which offers a beat-wise annotation and therefore allows for an easier interpretation of obtained attention outputs.

\section{Related Work}
\label{sec:recentwork}
This brief overview of related AF classification systems focuses on recent attempts to combine both CNNs and LSTMs for the application of ECG processing. The discussion of convolutional long short-term memory networks (ConvLSTMs) for other applications would go beyond the scope of this introduction. Still, it is worth mentioning that ConvLSTMs already were successfully applied in many other tasks such as sleep stage scoring on EEG data \cite{supratak2017}, weather forecasting \cite{Xingjian2015}, 
and gesture recognition (outperforming both plain CNNs and plain LSTMs) \cite{tsironi2016}. Finally, work in the area of recent attention mechanism and attention network variants and applications will be reviewed.

\subsection{ECG Rhythm Classification}
Following the analysis of a literature study by Jangra et al.~\cite{jangra2017}, established ECG classifiers typically include a preprocessing (e.g. removing noise and artifacts), a feature extraction, a feature space dimensionality reduction (e.g. performing independent component analysis), and a classifier module. The feature extraction part aims at describing the signal by morphological, temporal, and statistical features and is often based on cardiologists' expert knowledge. Typical features are, for instance, capturing information about the extrema of waveforms, RR intervals, QRS complex durations, and wavelet transforms \cite{jangra2017}. In their overview of state-of-the-art algorithms, Jangra et al. name, among others, feed forward networks \cite{martis2012}, probabilistic networks \cite{martis2012}, support vector machines \cite{martis2012}, 
fuzzy neural networks \cite{shyu2004}, 
radial basis function neural networks \cite{yang2010}, and
random forest algorithms \cite{emanet2009}. 

\paragraph{CinC 2017 challenge participants}
Some of the before mentioned algorithms did also participate in the CinC challenge 2017 (see \cite{clifford2017} for all 75 participating teams). Amongst the submissions are support vector machines \cite{behar2017}, random forest classifiers \cite{zabihi2017}, CNNs \cite{rubin2017}, RNNs \cite{teijeiro2017}, and also some ConvLSTMs \cite{alrahhal2018}, \cite{warrick2017}. One of the winning teams that reached an average F1 score of $ 0.83 $ were Teijeiro et al.~\cite{teijeiro2017}. They combined an LSTM network with a tree gradient boosting (XGBoost) classifier and utilized the Construe algorithm in order to find ``same features as used by cardiologists'' regarding morphology, rhythm and signal quality.

A slightly worse F1 score of $0.82$ (while using a less complex architecture, but processing the ECG records as logarithmic spectrograms) was achieved by Zihlmann et al.~\cite{zihlmann2017}. 
They combined a deep CNN with a multi-layer bidirectional LSTM for the temporal aggregation of features. The authors also compared the performances of LSTMs and temporal average pooling operations and showed that the LSTM slightly outperformed the pooling variant in case data augmentation was employed. Reviewing their experiments, Zihlmann et al. did not observe great performance differences between the two variants. Still, they formulated the hypothesis that the aggregated feature vector provided by an LSTM could potentially better preserve episodic phenomena than a simple pooling layer. Another team processed raw ECG data using a ConvLSTM with only one CNN layer and thereby obtained an F1 score of $0.80$ without requiring any pre-processing of the data \cite{warrick2017}.

Despite the comparably large amount of submitted RNN approaches, there were also teams applying pure CNNs or pure LSTMs only. The densely connected CNN of Rubin et al.~\cite{rubin2017}, for instance, processed time-frequency representations of the data and achieved an F1 score of $0.80$. 
The team additionally incorporated information about the signal quality to immediately classify some records as noisy and furthermore used an Ada-Boost classifier in a post-processing step. Another pure CNN architecture was the one-layer CNN architecture of Chandra et al.~\cite{chandra2017}, which processed heartbeat windows centered at detected R-peaks but only achieved an F1 score of $0.71$. 
A pure LSTM, on the contrary, was submitted by Maknicka et al.~\cite{maknickas2017}, who used pre-computed QRS features as input to a multi-layer LSTM and thereby achieved an F1 score of $0.78$. 

Observing that many approaches of the challenge used complex pre- or post-processing steps, this work aims at setting up a simpler architecture which only relies on CNN and LSTM modules and allows for the processing of raw ECG data. Since CNNs proved to be promising feature extractors for the given task, the focus of our work is the comparison of different temporal feature aggregation strategies (comparing particularly simple pooling layers with more complex LSTM modules). 

\subsection{Attention Mechanisms}

First neural network attention mechanisms were proposed for CNNs, being either applied in a post-processing step (after a networks training had finished) \cite{simonyan2013}, \cite{zhao2017} or integrated into the learning process (requiring the optimization of additional attention parameters) \cite{bahdanau2014}, \cite{schlemper2018}.  

In some of the first works, Zeiler et al.~\cite{zeiler2014} used deconvolutional networks to trace high neuron activations back to the image space. By visualizing resulting attention maps, the authors managed to identify input objects and patterns that were of highest relevance for the prediction of the network. In this work, we will study the attention visualization approach that was proposed by Zhou et al.~\cite{zhou2016} in 2016 and has subsequently been successfully applied by Wang et al.~\cite{wang2017} for the weakly supervised localization of thorax diseases in chest X-ray images. It was found that the proposed class activation map (CAM) concept provides a convenient way to highlight salient input regions as it only requires a few modifications of the standard CNN classifier part.

In another work, Fong et al.~\cite{fong2017} introduced model independent attention maps that are based on the computation of occlusion masks. By masking out most informative input pixels (with e.g. a constant value or noise), evidences that led to the original class decision are removed and networks consequently are fooled. Westhuizen et al.~\cite{westhuizen2017} recently applied a `zero perturbation' occlusion mask for heartbeat windows of the MIT-BIH database. It is assumed that for the rhythm classification task the expressiveness of resulting attention maps can be further improved by a more realistic perturbation variant. For this reason, the estimation of a deformation grid that applies a temporally varying shift to the ECG data will be introduced and explored in Sec.~\ref{sec:attention}.

Beside the idea of visualizing attention, there has been some further research on how to benefit from attention to improve the performance of networks. So-called attention networks are motivated by the human visual attention system and have before, for instance, successfully improved fine-grained object classifications \cite{zhao2017}. Some further networks were proposed that learned to apply suppression masks \cite{dahun2017}, to ``look and think twice'' \cite{cao2015}, or to ``pay more attention to attention'' \cite{zagoruyko2016}. This work studies attention gated CNNs that aim at incorporating the most contextually `useful' features of multiple network layers to more explicitly provide features at different resolutions. Those networks were introduced by Jetley et al.~\cite{jetley2018}, were extended by Oktay et al.~\cite{oktay2018} for the task of organ localization in abdominal CTs and also showed to improve ultrasound scan plane detections \cite{schlemper2018}. 

\chapter{Basic Concepts} 
\label{sec:basics}

After a short introduction to electrocardiograms and the diagnosis of atrial fibrillation, this section presents the basic concepts of first, CNNs and second, RNN architectures.

\section{Electrocardiogram Basics}
\label{sec:ecgbasics}

Since the recording of electrical heart activity conveniently visualizes depolarization and repolarization disorders of cardiac fibers, the detection of cardiac arrhythmias is commonly based on the examination of ECGs (where disorders are indicated by changes of the P, Q, R, S, and T wave amplitudes and intervals).

\begin{figure}
    \centering
    \begin{subfigure}[b]{0.45\textwidth}
        \includegraphics[width=\textwidth]{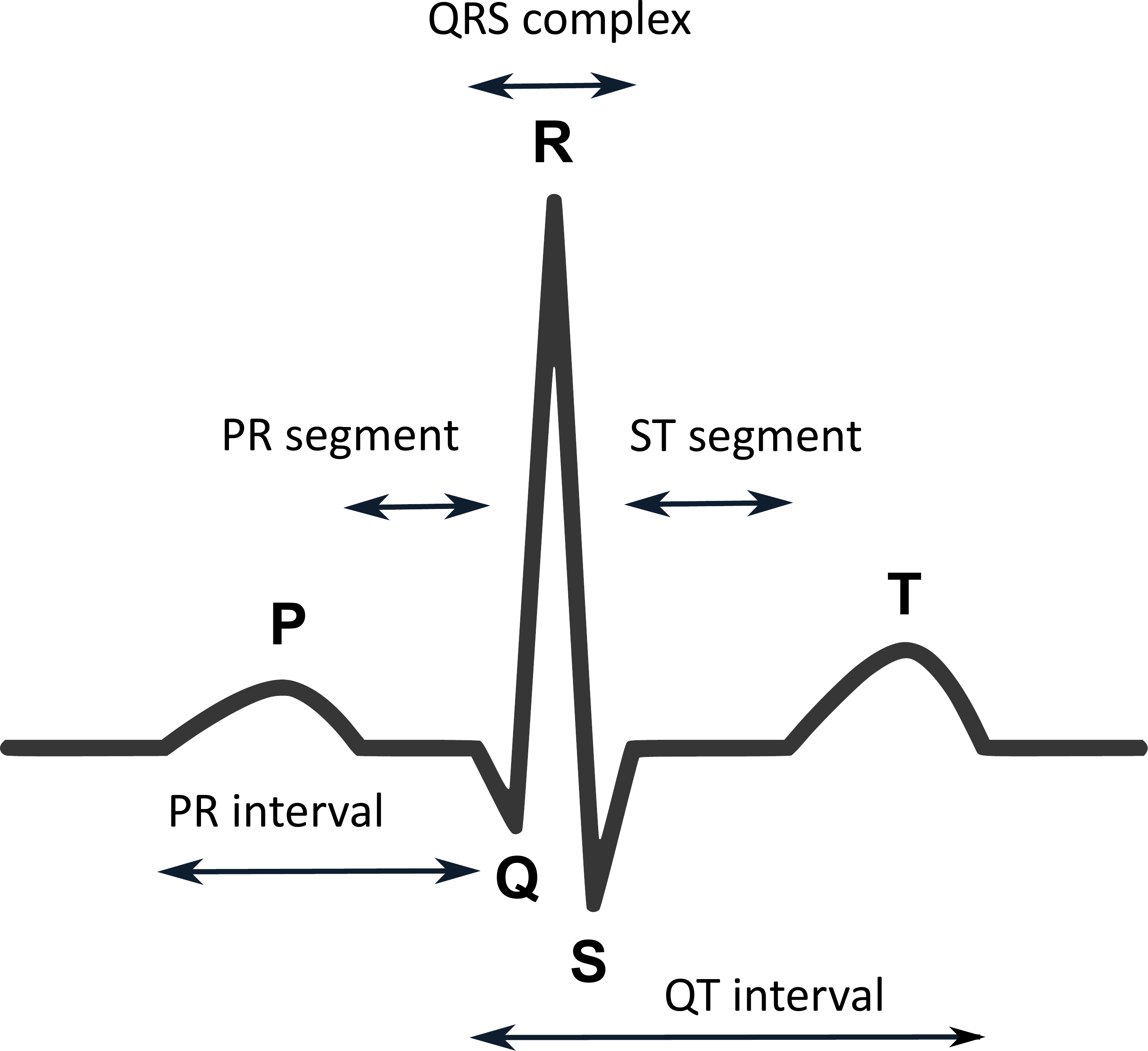}
        \caption{} 
        \label{fig:ecg0}
    \end{subfigure}
    \begin{subfigure}[b]{0.5\textwidth}
        \includegraphics[width=\textwidth]{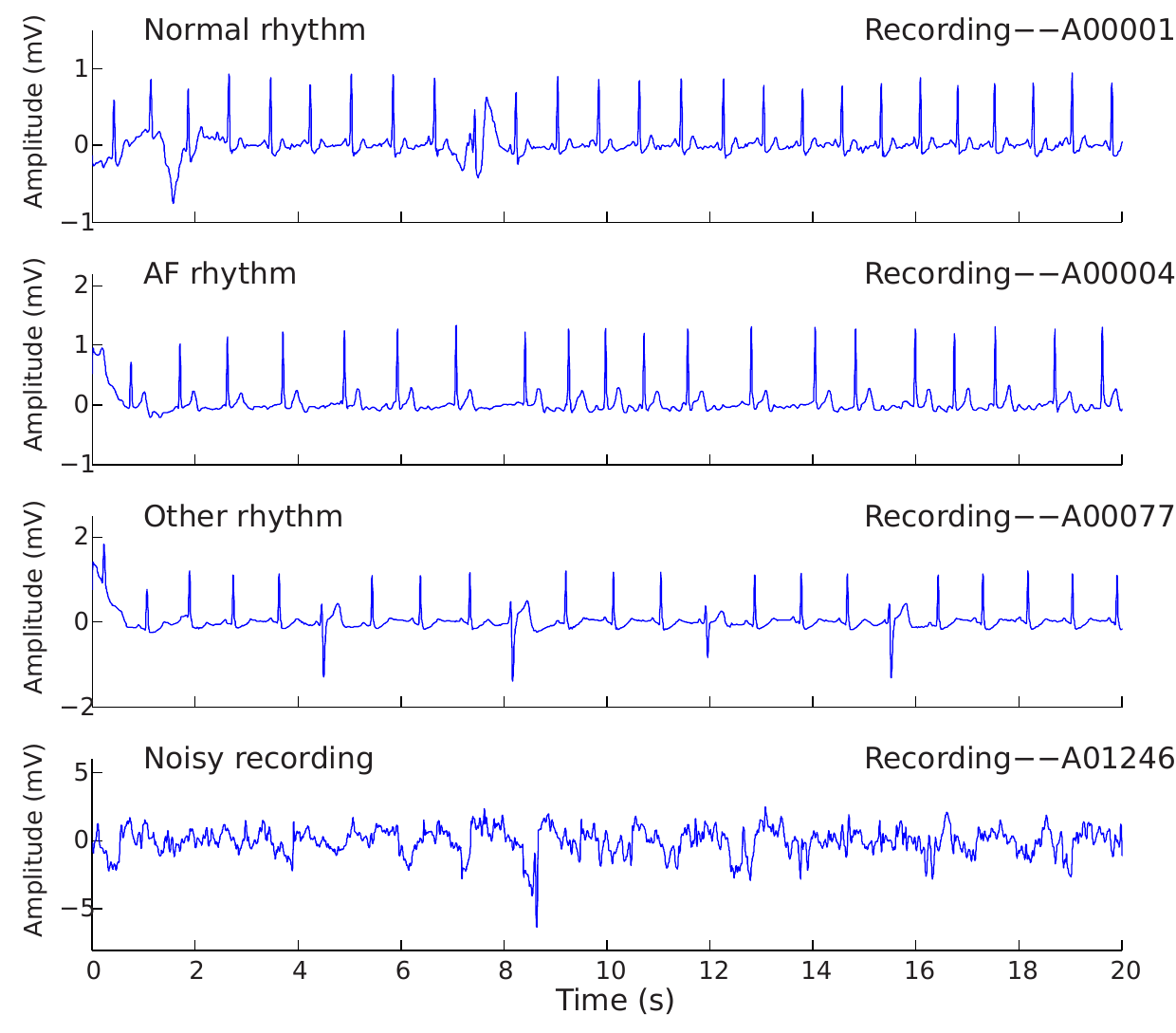}
        \caption{}
        \label{fig:ecg1}
    \end{subfigure}
    \caption{(a) Schema of an ECG curve and (b) example records of the CinC database. Source: \cite{clifford2017}.}
\end{figure}

Figure \ref{fig:ecg0} shows a diagram of a typical healthy heartbeat that represents a normal cardiac contraction cycle. Its shape originates from the electrical activity of the following processes: the depolarization of the atrial muscle fibers and pass through the atrioventricular (AV) node (P wave), the activation of the muscle fibers in the ventricles (QRS complex), the plateau of the ventricular action potential (ST), and the repolarization of the ventricles (T wave) \cite{selzer1970}. 

Regarding the classification task of the CinC challenge 2017, normal sinus rhythms (N), atrial fibrillation (AF), other rhythms (O), and signals that are too noisy to be classified ($\sim$) are to be distinguished (see Fig. \ref{fig:ecg1} for example signals). Cardiac rhythm changes (that can be of permanent or episodic nature) typically arise if the cardiac cycle of the sinus rhythm is disturbed by, for instance, abnormal electrical activations originating from the atria, the AV node, or the ventricles. 

\subsection{Atrial Fibrillation}

Atrial fibrillation is 
caused by chaotic electrical activations in the atria and 
results in irregular heart contractions that disturb the mechanical functions of the heart and also affect the whole cardiac-vascular system (potentially leading to heart failure, stroke, coronary artery disease and the risk of death \cite{goldberger2000}). The development of reliable AF detectors is therefore important but at the same time challenging due to episodic occurrences of AF, noisy clinical data, and the differentiation between AF and other arrhythmias which show similar ECG features (like irregular RR intervals).

\paragraph{AF related ECG appearances} 
In cases of AF, irregular RR intervals result from the fibrillation of the atria which excites the AV node at a very high rate. This in turn, causes the AV node to fire in a chaotic and highly irregular pattern and consequently leads to an irregular polarization of the ventricles. It furthermore affects the morphology of the sinus waveform in such a way that the `traditional' P wave is replaced with the appearance of low amplitude f waves. That is why common AF classification systems need to capture both abnormal morphologies and rhythm changes.

\begin{figure}
\centering
\includegraphics[width=1.0\linewidth]{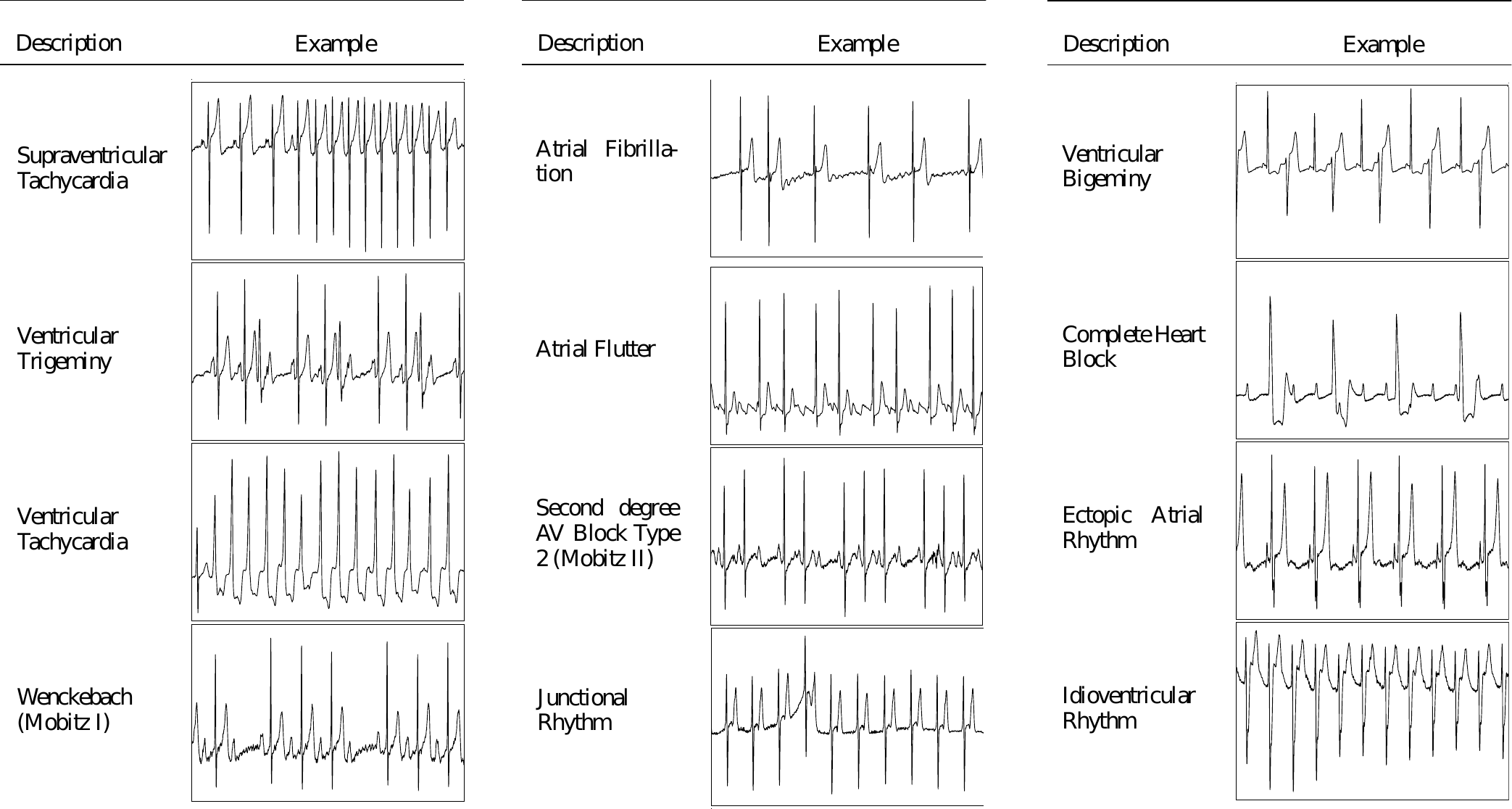}
\caption{ECG record examples showing the variety of common cardiac arrhythmias. Source: \cite{rajpurkar2017}.}
\label{fig:rhythm_examples}
\end{figure}

\paragraph{Class Other related ECG appearances}
As the class Other comprises various cardiac pathologies, possible appearances are more difficult to summarize than those of the AF class. Figure \ref{fig:rhythm_examples} shows a few examples of popular rhythm disorders that were presented in \cite{rajpurkar2017}. It is to be noted, that those examples were not extracted from the CinC database. Still, the selection of records conveniently shows the variety of other rhythm types and illustrates the challenge of representing all those abnormalities with a training set of less than ten thousand patients. Further challenges are introduced by the presence of noise and varying heart rates (with a rate of more than 100 beats per minute (bpm) being referred to as tachycardia and one of less than 60 bpm being referred to as bradycardia). As it can be seen in Fig.~\ref{fig:rhythm_examples}, class Other records show similar RR interval changes as AF records but often also exhibit morphology changes of e.g. the QRS complex. 

\section{Neural Network Basics}
\label{sec:networks}
In the following section, four commonly applied network architectures will be introduced for the task of ECG classification: convolutional neural networks (CNNs) \cite{lecun1998}, vanilla recurrent neural networks (RNNs) \cite{elman1990}, long short-term memory networks (LSTMs) \cite{hochreiter1997}, and gated recurrent unit networks (GRUs) \cite{cho2014}.

Throughout this work, a weakly supervised classification setup will be considered where each ECG record $X={x_{0},x_{2},...,x_{N-1}}$ of length $N$ is assigned with one target rhythm label $c \in {1,2,...,C}$. For this purpose, all network architectures will apply a final fully connected classifier layer with softmax activation to yield a pseudo-probability $\hat{y}_{c} \in [0,1]$ for each possible output class $c$. In the following, the network specific ways of extracting and processing class discriminative features of the input will be studied. 

\subsection{Convolutional Neural Networks}
\label{sec:methods:cnn}
Beside many successes in image processing tasks, CNNs were recently also applied to accurately classify temporal data. Two key aspects that make CNNs such powerful are location invariance and the composing of increasingly complex features. 

\paragraph{The concept of feature maps}
Location invariance is achieved by convolving input images (or signals) with shared filters which allows for the detection of same patterns at different locations of the input. It has been shown that first layers extract simple features like edges and subsequent layers gradually build up more complex features by combining the patterns of the preceding layer (building shapes from edges, object parts from shapes and finally entire objects from object parts). The number of patterns which can be detected by a layer depends on its number of channels since each channel learns one filter for the convolution. The channel-wise filter responses (feature detected or not) 
result from the convolution of a channel filter with the input of the layer and are stored in so-called feature maps.

More formally, the output activations of the $j$th channel for a given layer $l$ are the results of the convolution of an input activation $a_{in}$ with a shared filter $w$: 
\begin{equation}
a_{out, j}^{l} = f(\sum_{k=0}^{C_{in}^{l}-1}w_{j}^{l} * a_{in,k}^{l} +b_{j}^{l}).
\end{equation}
If the input $a_{in}$ consists of several channels, the convolution (denoted as $*$) is performed and summed up over all $C_{in}$ input channels. 
Commonly, a non-linearity $ f $ is afterward applied to the output to yield the output activations $a_{out, j}$. One example of such a non-linear function is the rectified linear unit (ReLU) which computes: 
\begin{equation}
ReLU(z_{j}^{l})=\max(0,z_{j}^{l}).
\end{equation}
\begin{figure}
\centering
\includegraphics[width=1.0\linewidth, trim={0 16cm 0 0}, clip]{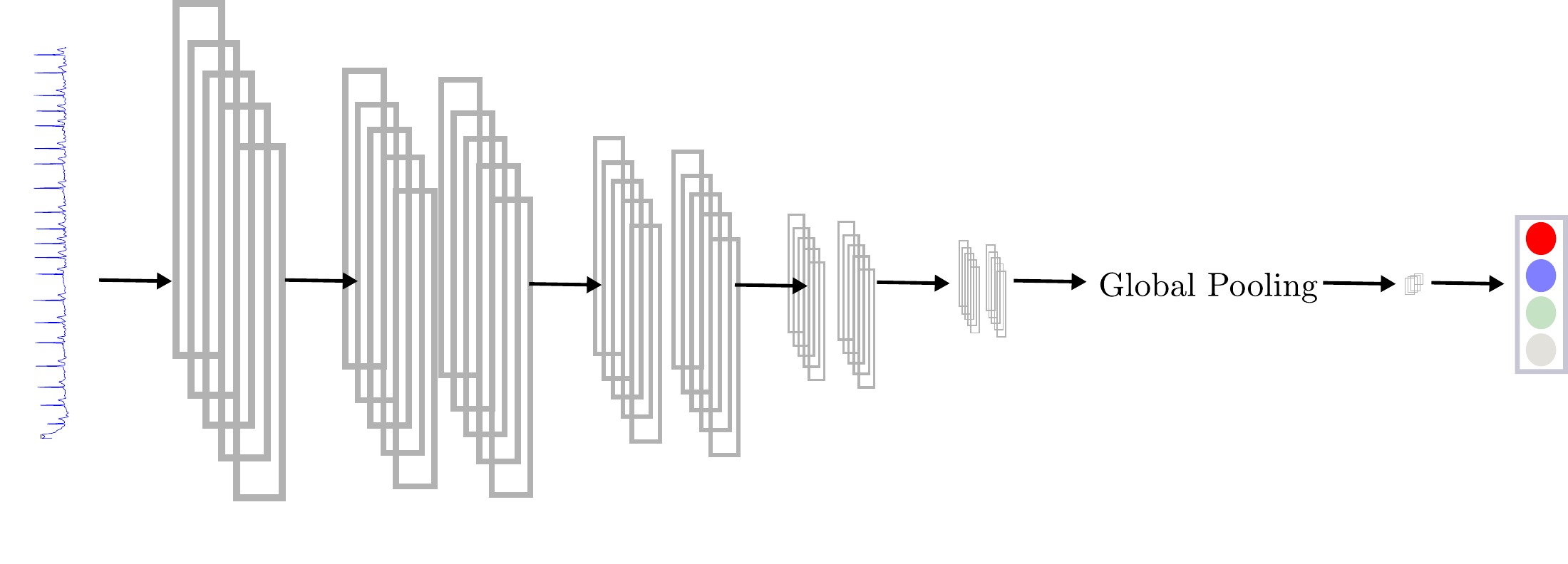}
\caption{Convolutional neural network architecture using global pooling for the channel-wise aggregation of features to obtain a fixed sized vector that can be passed to the classification layer.}
\label{fig:cnn_net}
\end{figure}

\paragraph{Global pooling for weakly supervised classification}
For global classification tasks, the output feature map of the last convolution layer needs to be reduced to a fixed size vector before being passed to a final classification layer (see Fig. \ref{fig:cnn_net}). For this task, a global pooling operation is applied (e.g. global maximum pooling (GMP) or global average pooling (GAP)).
Let $a_{out}^{L}(n)$ be the output activations of the last convolutional layer $L$ at temporal (or spatial) location $n$. To channel-wise aggregate features to one scalar, the global average pooling for each channel $j$ is defined as: 
\begin{equation}
gap_{j}^{L}= \frac{1}{N} \sum_{n=0}^{N-1} a_{out,j}^{L}(n).
\end{equation}
The resulting feature vector is subsequently connected to all $C$ neurons of the classification layer (with each connection holding one weight $w_{j,c}$). The linear classification layer is then computing output scores $ s_{c} $ for each neuron $c$:
\begin{equation}
s_{c}=\sum_{k=0}^{C^{L}_{out}-1} w_{k,c} gap_{k}^{L} + b^{c},
\end{equation}
with $C^{L}_{out}-1$ again being the number of channels in the last convolutional layer. Finally, a softmax function can be applied to squash the output scores into the range $[0,1]$, summing up to 1 over all $C$ classes:
\begin{equation}
softmax(s_{c}) = \frac{\exp(s_{c})}{\sum_{c=1}^{C}\exp(s_{c})}.
\end{equation}
Both shared weights and intermediate pooling layers have a positive effect on the generalization abilities of a network. Since pooling layers discard information about location of patterns, they introduce both translation as well as some scaling invariance. 

\subsection{Recurrent Neural Networks, Long Short-Term Memory Networks, and Gated Recurrent Unit Networks}
\label{sec:methods:lstm}
\begin{figure}
    \centering
     \begin{subfigure}[b]{0.49\textwidth}
	\includegraphics[width=1.0\linewidth, trim={0 0 0 0}, clip]{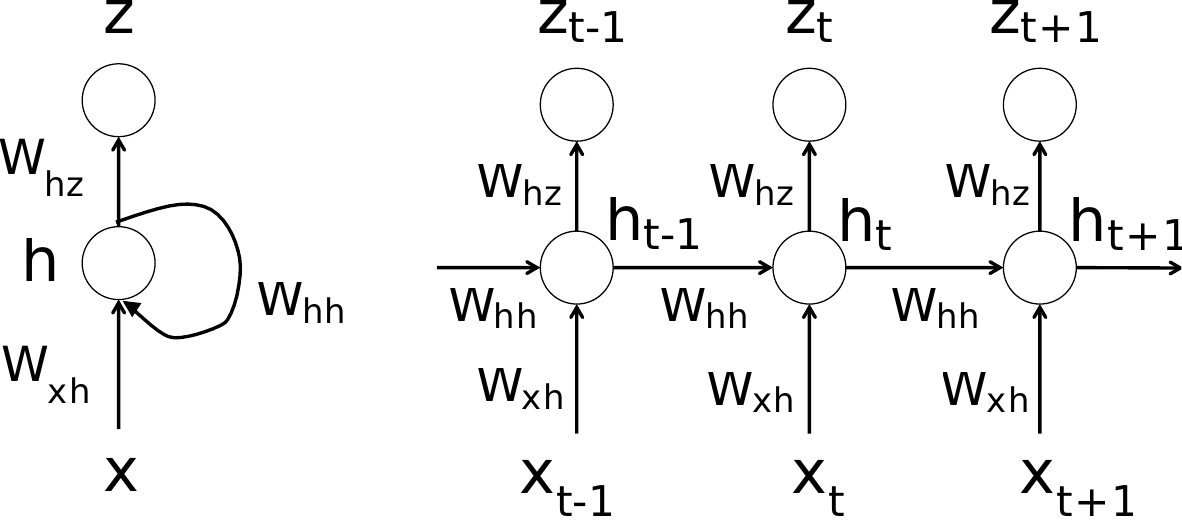}
	\caption{}
	\label{fig:unrolledrnn}
     \end{subfigure}
    \begin{subfigure}[b]{0.49\textwidth}
    \includegraphics[width=1.0\linewidth, trim={0 0.5cm 0 0}, clip]{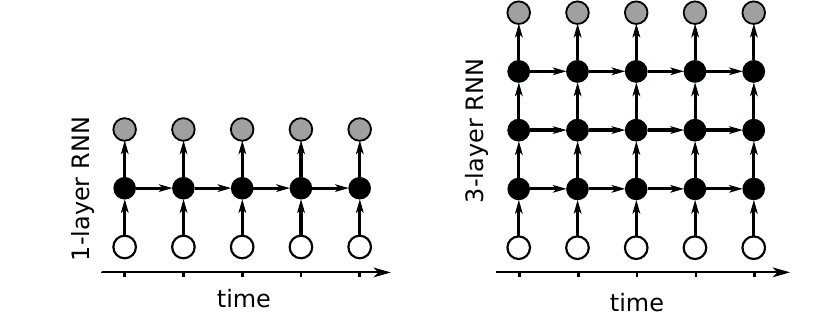}
    \caption{}
    \label{fig:multilayerRNN}
    \end{subfigure}
	\caption{(a) RNN with a recurrent connection visualized as recursive graph (left) and as unrolled sequential time graph (right). Source: \cite{chen2016}. (b) Many-to-many RNN performing a sequence-to-sequence prediction (left). Multi-layer variant where each RNN layer receives the hidden state sequence of the previous layer as input (right). Source: \cite{hermans2013}.}
\end{figure}

Recurrent neural networks have been introduced for the analysis of data that is changing over time. Popular applications are, for instance, speech recognition \cite{sainath2015}, image captioning \cite{chen2016}, or character prediction for the impressive generation of (almost compiling) source code \cite{karpathy2015}). An idea that comes along with the processing of time series is that the output of a neuron should not only depend on a given time step input but rather on information of the (entire) past. To allow the network to memorize and to access input histories, so-called recurrent connections are inserted. Figure \ref{fig:unrolledrnn} illustrates such a residual connection which is basically a simple feedback loops from a cell to itself. 
This section will first study the realization of internal memories for standard (vanilla) RNNs and will afterward present the two most popular RNN variants that are LSTMs and gated recurrent unit networks (GRUs). 

\paragraph{Vanilla recurrent neural networks}

The loop that is introduced by a recurrent connection is often visualized in a `time-unrolled' way. Figure \ref{fig:unrolledrnn} depicts the resulting sequential graph as a chain of repeated cell modules (where the number of repetitions corresponds to the number of inputs in the input sequence). Regarding vanilla RNNs, those modules are simply repeated tanh layers, where a module output at time $ t $ is defined as \cite{olah2015}:
\begin{equation}
h_{t}=\tanh(W_{hh}h_{t-1}+W_{xh}x_{t}), 
\end{equation}
using the tanh function:
\begin{equation}
\tanh(x) = \frac{e^x - e^{-x}} {e^x + e^{-x}}.
\end{equation}
Thus, the cell holds a weight for both, outputs of the last time step hidden state $h_{t-1}$ ($W_{hh}$) and new input $x_{t}$ ($W_{xh}$). By updating the parameters of the weight matrices, the cell learns to memorize information of the past and to decide which information to keep or to overwrite by new inputs.

Depending on the application, RNNs can, among others, be applied for sequence-to-sequence predictions (many-to-many RNNs) or, as in case of ECG rhythm classification, for the predictions of global class labels after the processing of all input samples has finished (many-to-one RNNs). For both variants, the hidden states serve as hidden representation of the inputs and are passed to the classification layer for the class prediction. Softmax scores at time step $t$ are computed as:
\begin{equation}
z_{t} = softmax(W_{hz}h_{t}+b_{z}),
\end{equation}
with $W_{hz}$, $b_{z}$, as usual, being the weights and the bias of the output layer.

As it is the case for multi-layer CNNs, it is possible to stack multiple layers of RNNs in order to built up temporal features hierarchically. For multi-layer RNNs, each RNN layer receives the hidden state sequence of the previous layer as input (see Fig.~\ref{fig:multilayerRNN}). Hermans et al.~\cite{hermans2013} argue that for the application of speech recognition such a hierarchy corresponds to a processing of time series at several time scales (building up features of words, phrases, sentences, and finally, full conversations at the highest level).  

Looking at the unrolled RNN visualization of Fig.~\ref{fig:unrolledrnn}, it becomes apparent that the weights $W_{xh}, W_{hh}$, and $W_{hz}$ are shared across all time steps. Given the global classification approach, the class prediction is performed after the processing of the entire sequence (that is after the last time step T). As for CNNs, the loss (here the cross-entropy loss) is first computed in a forward pass. In a subsequent backward pass, the partial error derivatives w.r.t. the network parameters (weights and biases) are then computed for the Stochastic Gradient Descent (SGD) update. 
Given that the RNN output $z_{T}$ depends on all previous time steps, backpropagation for RNNs is backpropagation through time (BPTT) and associated with a recursive application of the chain rule. A detailed derivation of the BPTT formulas can be found in a publication of Chen et al.~\cite{chen2016}. Here, only one formula is given in order to hint the problem that arises for the processing of very long sequences. Following Chen et al., the derivative for the weight $W_{hh}$ (when considering a cross-entropy loss) is given by \cite{chen2016}:

\begin{equation}
\frac{\partial L}{\partial W_{hh}} = \sum_{t}\sum_{k=1}^{t+1}\frac{\partial L(t+1)}{\partial z_{t+1}}\frac{\partial z_{t+1}}{\partial h_{t+1}} \frac{\partial h_{t+1}}{\partial h_{k}} \frac{\partial h_{k}}{\partial W_{hh}}.
\end{equation}
It can be deducted from this formula that gradients are aggregated over the whole time sequence length $T$. As this aggregation can include the multiplication of a large number of small factors, gradients are at risk to get increasingly small.

\begin{figure}
\centering
\includegraphics[width=0.5\linewidth]{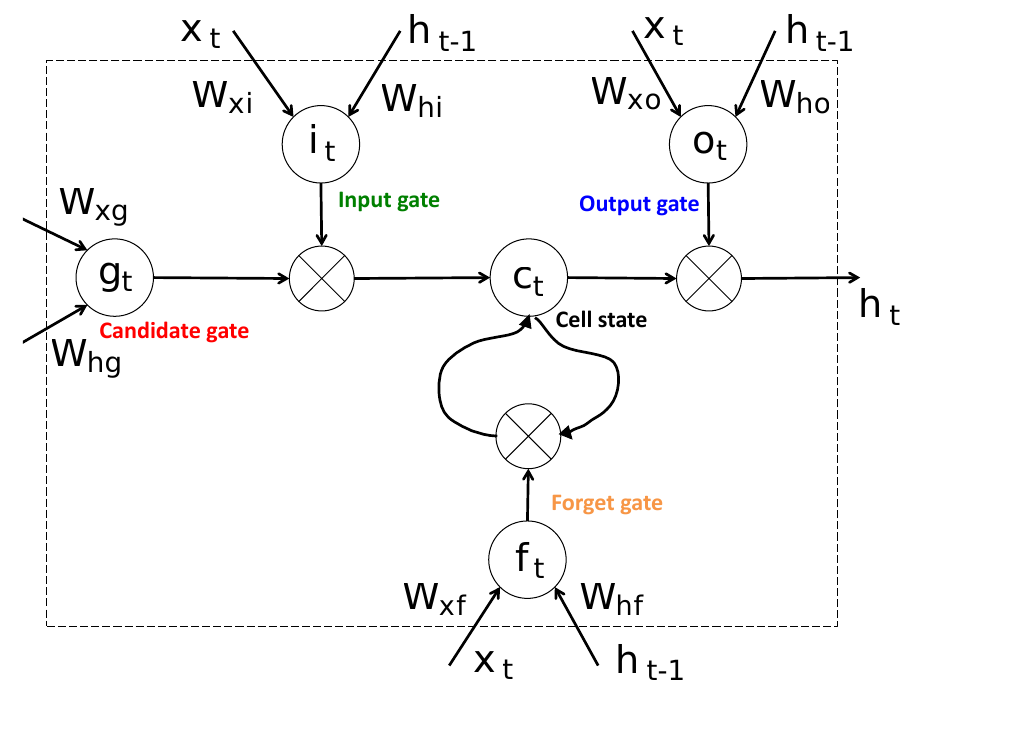}
\caption{LSTM cell structure showing the relations between the hidden state $h_{t}$, the cell state $c_{t}$ and the four cell gates which determine the accessing of the cell memory. Source: \cite{chen2016}.}
\label{fig:lstmcell}
\end{figure}

\paragraph{Long short-term memory networks}
Given the fact that this popular `vanishing gradient' problem can impede the learning process severely, LSTMs were proposed to better control the gradient flow. The LSTM cell formulation is for this reason extended by the definition of gate units and a memory state \cite{chen2016}. Figure \ref{fig:lstmcell} illustrates the structure of such an LSTM unit including the input, output, candidate, and forget gates that are known to enable the memory storage for long sequences (by e.g. keeping the cell from being overwhelmed by irrelevant inputs). For each sample $x_{t}$ of the input sequence an LSTM unit computes the following functions at time step $t$ \cite{paszke2017}:

Input gate: 
\begin{equation}
i_{t}=\sigma(W_{xi}x_{t}+b_{xi}+W_{hi}h_{t-1}+b_{hi}) \label{eq:lstmfirst}
\end{equation}

Forget gate: 
\begin{equation}
f_{t}=\sigma(W_{xf}x_{t}+b_{xf}+W_{hf}h_{t-1}+b_{hf}) 
\end{equation}
Output gate: 
\begin{equation}
o_{t}=\sigma(W_{xo}x_{t}+b_{xo}+W_{ho}h_{t-1}+b_{ho}) 
\end{equation}
Candidate gate: 
\begin{equation}
g_{t}=\tanh(W_{xg}x_{t}+b_{xg}+W_{hg}h_{t-1}+b_{hg})
\end{equation}
Hidden state:
\begin{equation}
h_{t}=o_{t}\tanh(c_{t}) \label{eq:lstmlast}
\end{equation}

Cell state:
\begin{equation}
c_{t}=f_{t}c_{t-1}+i_{t}g_{t}
\end{equation}
where $h_{t-1}$ and $c_{t-1}$ are the hidden state and the cell state of the previous time step (or the initial states at time step $t=0$). The gates apply the sigmoid function $\sigma$:
\begin{equation}
\sigma(x) = \frac{1}{1 + \exp(-x)}
\end{equation}
to compute scalars in the range $[0,1]$ (to either let all information pass by opening the gates which is corresponding to a value of 1 or to not let any information pass with a value of 0). 

While the cell state $c_{t}$ is considered to be a long-term memory, the hidden state $h_{t}$ rather represents the working memory which focuses on immediately useful information of the long-term memory \cite{chen2016}. The hidden state is therefore claimed to be a sharped version of the cell state (since it results from a multiplication of the cell state with the output gate, which limits the information that is passed on) \cite{chen2017}. During the training process, the LSTM adapts the weights $W$ that are associated with all gates to learn which information to remember, to update, and to pay attention to. Short, it learns a good feature representation $h_{t}$ of the data, which finally can be used as input for the classification layer.

\begin{figure}
\centering
\includegraphics[width=1.0\linewidth]{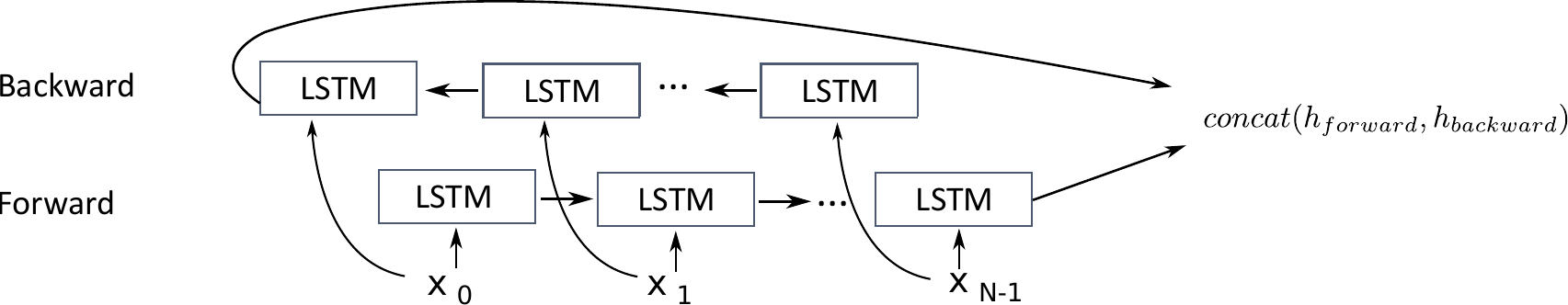}
\caption{Bidirectional LSTM concatenating the last hidden states of the forward and the backward pass as input to the classification layer.}
\label{fig:bidirectional_lstm}
\end{figure}

\paragraph{Bidirectional long short-term memory networks} 
Bidirectional LSTMs can be beneficial in setups where an output at a given time step not only depends on the past but also on the future information. The idea is to basically train two independent LSTM units, one `forward' LSTM that processes the input sequence with regular ($ 0,..,N-1 $) time order and a second `backward' LSTM with reversed time order ($ N-1,..,0 $) processing. To combine information of both directions, the forward and the backward hidden states are concatenated at each time step. For many-to-one LSTMs, commonly, the hidden states $h_{N-1}$ and $h_{0}$ are concatenated for the forward and the backward LSTM, respectively (see Fig.~\ref{fig:bidirectional_lstm}).

\paragraph{Gated recurrent unit neural networks}
GRUs are a simplified LSTM variant using a merged formulation of the cell and the hidden state as well as a merged formulation of the forget and the input gate into a single update gate (resulting in a total number of three gates). For each element in the input sequence, a GRU unit computes the following functions \cite{paszke2017}:

Reset gate:
\begin{equation}
r_{t} = \sigma(W_{ir} x_t + b_{ir} + W_{hr} h_{t-1} + b_{hr})
\end{equation}
Update gate:
\begin{equation}
z_t = \sigma(W_{iz} x_{t} + b_{iz} + W_{hz} h_{t-1} + b_{hz})
\end{equation}
New gate:
\begin{equation}
n_{t} = \tanh(W_{in} x_{t} + b_{in} + r_{t} (W_{hn} h_{t-1}+ b_{hn})) 
\end{equation}
Hidden state:
\begin{equation}
h_{t} = (1 - z_{t}) n_{t} + z_{t} h_{t-1}.
\end{equation}

\chapter{Methods}

As already discussed in Sec.~\ref{sec:recentwork}, many researchers are interested in combining the benefits of CNN and LSTM architectures. They suggest that using CNN modules for the hierarchical feature extraction task and a subsequent application of LSTM layers can help to capture temporal long-range dependencies in the feature sequence. Following the ConvLSTM motivation of Sainath et al.~\cite{sainath2015}, ``CNNs are good at reducing frequency variations, LSTMs are good at temporal modeling, and deep neural networks are appropriate for mapping features to a more separable space''. After explaining the general ConvLSTM setup used in this work, two attention mechanisms that can be applied for CNN architectures will be introduced. In addition, a network independent variant will be presented, that yields attention maps by computing perturbation masks for network inputs. 

\section{Convolutional Long Short-Term Memory Networks}
\label{sec:methods:cnnlstm}

\begin{figure}
\centering
\includegraphics[width=1.0\linewidth]{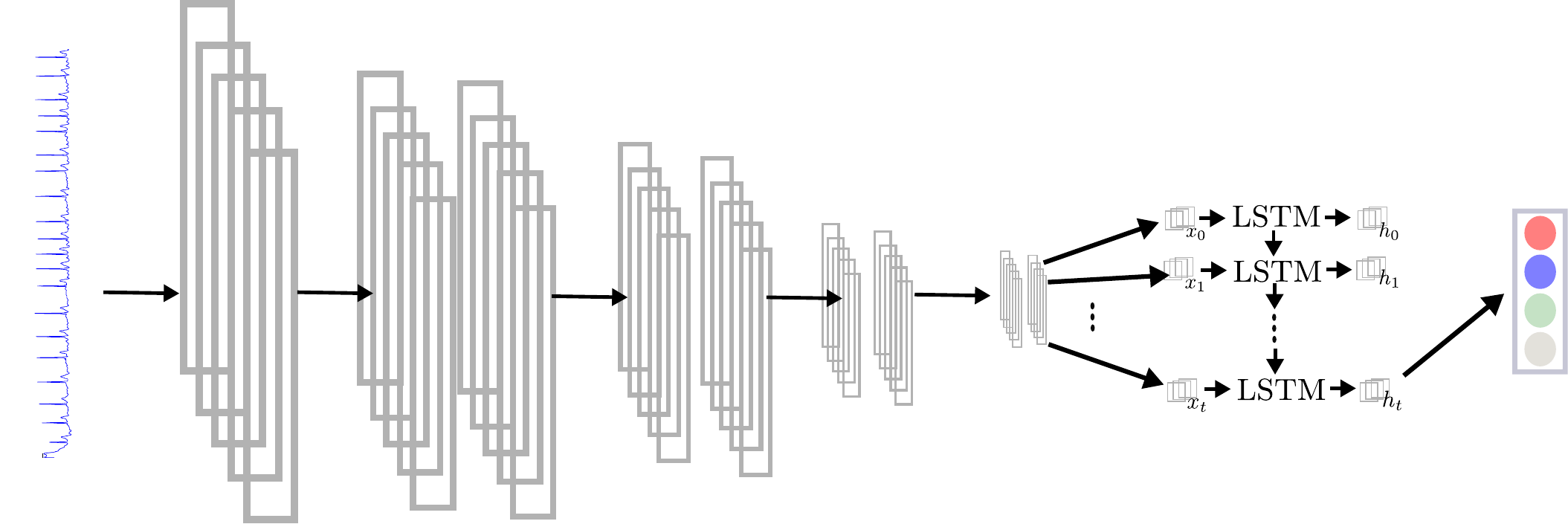}
\caption{ConvLSTM example setup using many-to-one inference. CNN feature vectors are extracted for each temporal sample (where the dimension of the feature vector corresponds to the amount of channels in the feature map) and are processed as a sequence by the LSTM. The LSTM is returning one hidden state at each time step and passes the last hidden state to the classification layer.}
\label{fig:cnn_lstm}
\end{figure}

Figure \ref{fig:cnn_lstm} shows the general setup of the ConvLSTM architecture used throughout this work. Feature vectors were extracted for each temporal location of the last CNN layer 
to define the input sequence for the LSTM module. The number of feature channels thereby defined the size of the feature vectors. 
In this work, we aimed at a simple data presentation and therefore proposed to present raw ECG data to the CNN module. However, it is also possible to previously transform the data into other representations (like logarithmic spectrograms \cite{zihlmann2017}). 

The task of the CNN is to extract a sequence of high-level features that can be easier processed by the LSTM than the raw ECG data representation. 
The deeper the CNN network (and the more pooling layers are applied), the smaller the feature map of the last layer becomes. If an LSTM would be applied on the raw ECG data instead, the parameter updates would require the backpropagation through all time steps (that are up to 18300 samples for the CinC challenge records). That is why a processing of raw data is often unfeasible for long sequences and the preceding feature sequence extraction of the CNN can help to reduce the number of time steps for the error propagation.

\newpage

Given the global classification task, in this work, a many-to-one LSTM is proposed. The many-to-one LSTM computes one hidden state for each input time step but only passes the last hidden state to the classification layer (see Fig.~\ref{fig:cnn_lstm}). It is assumed that the last hidden state incorporates memory information about the whole input sequence and implicitly represents those input features that were detected earlier in the sequence. Some recent works attempted to improve the incorporation of intermediate time step outputs by computing attention weighted combinations of hidden states (see for instance \cite{bahdanau2014}).

\section{Attention Mechanisms}
\label{sec:attention}
So far, there has been a range of attempts to visualize the internal processes of neural network models. In this section, we will first study the simple concept of class activation maps (CAMs) \cite{zhou2016} which provide heat maps that localize most important samples for the class decision. 
The related approach of attention gated CNNs \cite{schlemper2018} furthermore uses internal attention parameters to actively highlight or suppress information during the training process. This section afterward concludes with a presentation of perturbation masks, which are not restricted to the application of CNNs and derive attention maps by manipulating input data.

\subsection{Class Activation Maps}
\label{sec:methods:cam}
\begin{figure}
\centering
\includegraphics[width=0.9\linewidth]{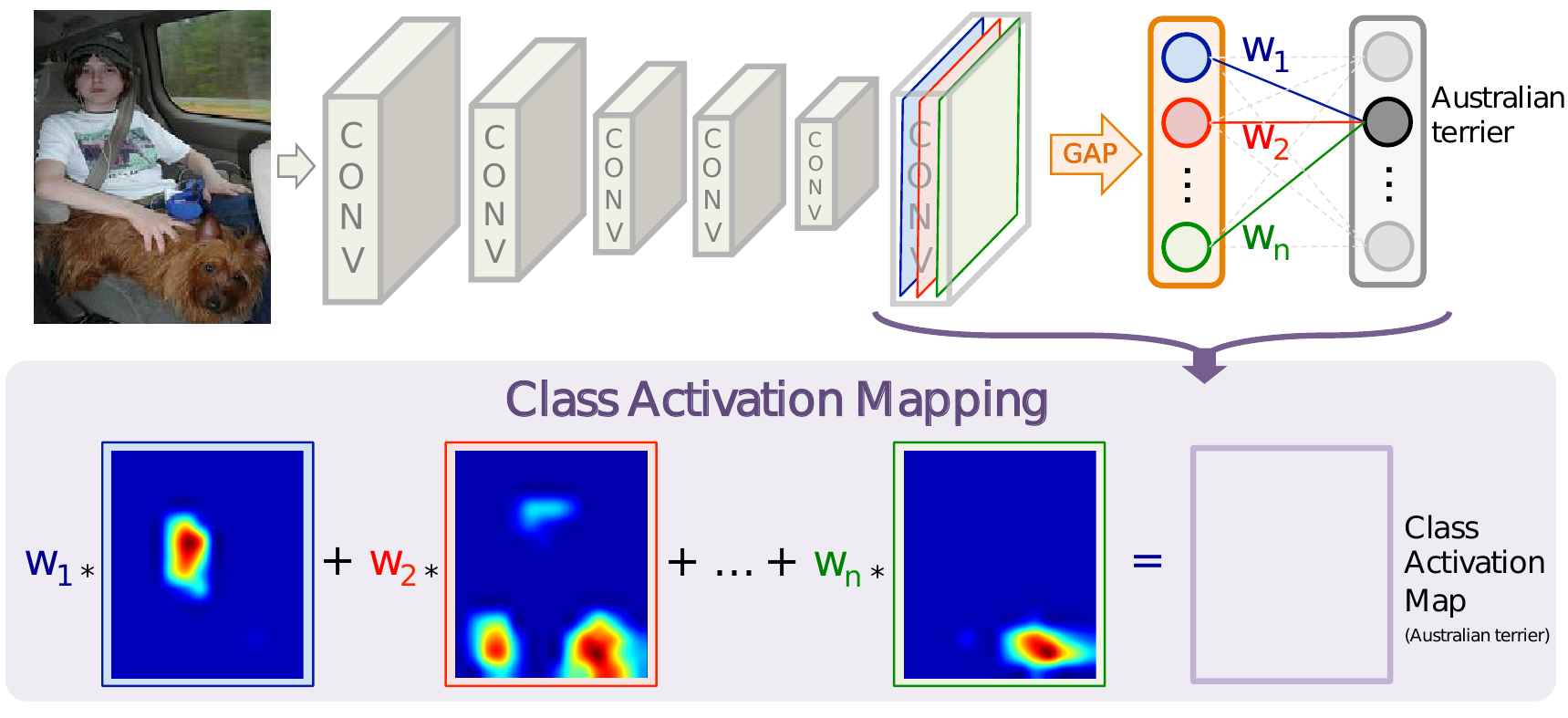}
\caption{Class activation map computation for the task of weakly supervised object localization. CAMs are computed as weighted linear sum of upsampled feature maps and weights that connect the globally pooled feature vector to the classification layer. Resulting attention maps can be interpreted as confidence maps that highlight most informative regions for the given class prediction. Source: \cite{zhou2016}.}
\label{fig:CamNet}
\end{figure}

Figure \ref{fig:CamNet} illustrates the basic architecture of Zhou's et al.~\cite{zhou2016} saliency visualization module, which requires
the insertion of a global pooling layer and a fully connected output layer after the last convolutional layer of a given CNN network. As discussed in Sec.~\ref{sec:methods:cnn}, each channel of a feature map within the network shows the presence or absence of one particular pattern (with patterns getting more and more complex with increasing network depth). After the CNN layers have completed the feature extraction part, the fully connected classification layer is supposed to learn which of the detected patterns are discriminative for each of the possible output classes. 

Fig.~\ref{fig:CamNet} demonstrates this concept for a simple example input. If, for instance, the pattern `dog nose' was detected in a given image (the associated feature map is colored in green), the output neuron that represents the class `Australien terrier' will get assigned a high weight $w_{n}$ during the training process. The feature `human face' (represented by the blue colored feature map), in contrast, will probably get assigned a weight $w_{1}$ close to zero. The global pooling layer is finally needed to reduce the spatial dimensions of each feature map channel to a scalar value that can afterward be passed to the classification layer (as it has been earlier discussed in Sec.~\ref{sec:methods:cnn}). The most commonly applied pooling variants are global average pooling, global max pooling, and log-sum-exp (LSE) pooling \cite{wang2017}.  

Even though the discarding of location information of detected patterns is supposed to not significantly damage the global classification performance, location information is often desired 
for weakly supervised localization tasks. That is why Zhou et al. proposed to recover location information by the computation of class activation maps. They are doing so by firstly upsampling the last layer feature maps, secondly extracting the weights that are connecting each pooled scalar to a given output neuron and thirdly computing the weighted linear combination of weights and upsampled maps. 
More formally, the class activation map for output neuron $c$ can be computed as weighted linear sum:
\begin{equation}
CAM_{c}=\sum_{k=0}^{C^{L}_{out}-1}w_{k,c}A_{out,k}^{L}, \label{eq:cam}
\end{equation}
where $A_{out,k}^{L}$ is the upsampled last layer feature map of channel $k$ and $w_{k,c}$ the weight connecting the pooled output of the given channel with the output neuron $c$. The result of this computation is a likelihood confidence map that allows to localize discriminative input regions for a given class prediction. Due to the required upsampling step, CAMs often provide rather rough localization of salient patterns (where the resolution of the map again depends on the parameterisation of the CNN).

\subsection{Attention Gates} 
\label{sec:methods:attentiongates}
\begin{figure}
\centering
\includegraphics[width=1.0\linewidth]{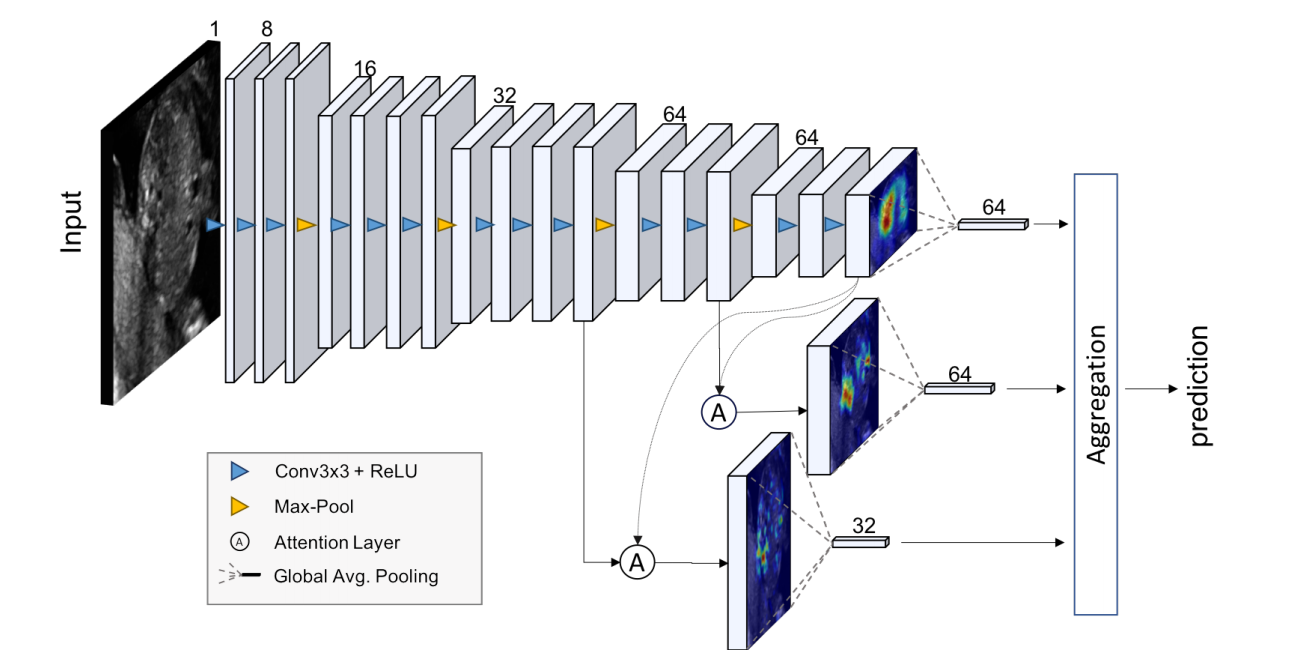}
\caption{Attention gate network introduced for the application of ultrasound scan plane detection. While the upper path shows the before described CNN setup with a global pooling layer, 
the network also comprises two attention paths. In those paths, intermediate feature maps are weighted according to attention weights that are defined by similarity scores to the last layer feature map. Again, the gated attention maps of both paths are globally pooled in order to provide feature vectors that are then, together with the upper layers feature vector, passed to the classification layer. Source: \cite{schlemper2018}.}
\label{fig:attentiongatecnn}
\end{figure}

Schlemper et al.~\cite{schlemper2018} argue that the application of global pooling layers forces CNN networks to extract the most salient features only on a very global level and that more local information should be preserved. By inserting attention gates at multiple CNN layers, they aim at incorporating salient information of different scales into the classification. 

Figure \ref{fig:attentiongatecnn} shows an example setup of their attention gated CNN architecture. While the upper network path represents a standard CNN applying a global pooling after the last convolutional layer, the network holds two additional attention paths. For each attention path, Schlemper et al. first extract an intermediate feature map, then weight the map according to a gating grid (that assigns one attention weight for each pixel location) and subsequently perform a global pooling to reduce the resulting gated attention map to a vector. The obtained vectors of all paths are afterward passed to the classification layer (e.g. using a concatenation strategy, where all vectors are concatenated before being passed to the last layer). 
To obtain the attention weights for each spatial location of a given intermediate feature map, Schlemper et al. compute compatibility scores to a global context grid. In their work, this global context grid is simply the feature map of the last layer (before the global pooling is applied). Attention coefficients are then defined by the following additive attention formula \cite{schlemper2018}:
\begin{equation}
\alpha_{i}^{l}=\sigma_{2}(\psi^{T}(\sigma_{1}(W_{x}^{T}x_{i}^{l}+W_{c}^{T}c_{i}+b_{c}))+b_{\phi}), \label{eq:attentiongate}
\end{equation}
with $W_{x}$, $W_{c}$, $\psi$, $b_{c}$, and $b_{\phi}$ being learnable parameters (implemented as $1\times 1$ convolutions), $\sigma_{2}$ the sigmoid function and $c_{i}$ a global context (gating) vector of the grid extracted at pixel location $i$. The motivation behind this complex similarity measure is the learning of a nonlinear, expressive relation between the intermediate and the last layer feature maps \cite{schlemper2018}. The actual `gating' is then a simple element-wise multiplication of feature maps and attention coefficients. For each layer $l$ it is defined as: 
\begin{equation}
\hat{x}_{i,k}^{l}=x_{i,k}^{l}\cdot\alpha_{i}^{l},
\end{equation}
where $\alpha_{i}^{l}$ is a scalar value for each pixel vector $x_{i,k}^{l}$ (which is shared over all feature map channels $k$). Finally, the weighted feature map of a given layer $l$ is pooled to a single output vector by simply summing all feature vectors over all pixel locations: 
\begin{equation}
g^{l}=\sum_{i=1}^{n} \hat{x}_{i}.
\end{equation}
Beside concatenating all attention gated vectors $g_{l}$, it is also possible to train fully connected layers for each of the paths independently. The output scores of all classification layers can then be combined utilizing e.g. an average or a maximum voting.

\subsection{Perturbation Masks}
\label{sec:methods:occlusionmask}

So far, only network architecture dependent attention visualization techniques were discussed that require the extraction of some internal parameters or even introduce additional parameters. Fong et al.~\cite{fong2017}, on the contrary, recently introduced a more general solution that can be applied to any model after training has finished. Their attention approach is inspired by the idea of a `deletion game' that progressively takes evidence from the input by perturbing salient regions (using different perturbation types like blur, constant occlusion values, or noise) in order to drop the confidence scores of an initial class prediction. Resulting `occlusion masks' identify those input pixels that had the highest impact on a model prediction and can therefore be considered as (inverse) attention maps. 

In order to find an occlusion mask $m$ that leads to a maximal drop of the initial prediction score, Fong et al. formulate an stochastic gradient descent optimization task. The objective function searches for a sparse and smooth mask (with $m_{i} \in [0,1]$) and is defined as \cite{fong2017}: 
\begin{equation}
arg\,min_{m} \lambda_{1} ||1-m||_{1}+\lambda_{2} \sum ||\nabla m||_{\beta}^{\beta}+s_{c}(\phi(x;M)). \label{eq:occlusionmask}
\end{equation}
In this formula, the first term minimizes the region masked, the second term the total variation and the third term the softmax score of the predicted class $c$ given the perturbed input. The parameters $\lambda_{1}$ and $\lambda_{2}$ are weights that define the influence of the L1 norm and total variation terms. To avoid over-fitting and the attraction to artifacts, a mask of low resolution is learned that is afterward upsampled to match the input image size (where the upsampled mask is denoted by $M$). The perturbation of the image $x$ with a constant value $k$ is computed as:
\begin{equation}
\phi(x;M) = M \odot x + k(1-m).
\end{equation}
While a mask value of 1 is not applying any perturbing, values close to zero replace the original input by the occlusion value completely (details about alternative perturbation types can be found in \cite{fong2017}). Given that low mask values correspond to high importances of a particular input region, attention heat maps can finally be computed as the normalized inverse of the occlusion masks.

\begin{figure}
\centering
\includegraphics[width=1.0\linewidth, trim={0, 3.5cm, 0, 0}, clip]{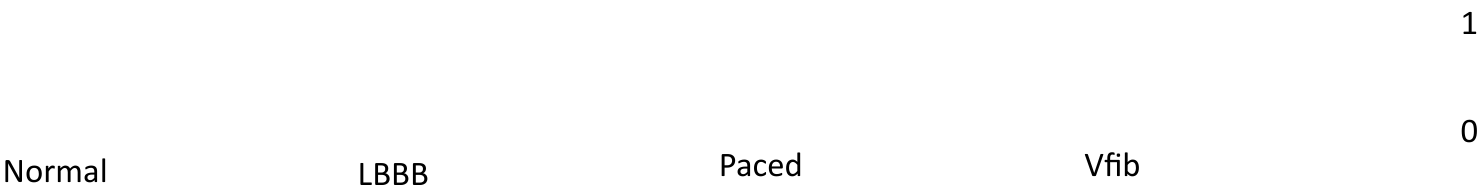}
\caption{Occlusion mask examples that have been reported for the task of MIT-BIH heartbeat classification (showing from left to right a normal beat, a left bundle branch block beat (LBBB), a paced beat and a ventricular fibrillation beat (Vfib)). In this setup, zero perturbation was applied to drop the sample amplitudes and resulted in masks that highlight class relevant patterns (the authors name e.g. a wider QRS complex for the LBBB beat and a lack of Q through for Vfib). Source: \cite{westhuizen2017}.}
\label{fig:westhuizen}
\end{figure}

Westhuizen et al.~\cite{westhuizen2017} recently studied the computation of occlusion masks for the heartbeat classification of MIT-BIH data and showed that perturbations with constant zero could yield meaningful attention maps (see Fig.~\ref{fig:westhuizen}). For the application of ECG rhythm classification, however, dropping sample amplitudes appear less useful. In order to obtain alternative, more realistic distortions we therefore propose the optimization of a shift deformation grid. Deriving attention maps from shift computations mainly focuses on the identification of abnormal, temporal features with, for instance, irregular RR interval. As discussed earlier, those rhythm irregularities can be primarily found in records belonging to the CinC classes AF or Other (when showing e.g. premature beats). Since the network is expected to potentially switch to class prediction Normal in case RR intervals are becoming increasingly regular (by shifting samples of sections where irregular intervals are observed), regions of large shifts are likely localizing those beats of strongest rhythm irregularity. For the experiments of this work, the objective function of Fong et al. will be only slightly adopted by defining $m$ as a perturbation grid instead of an occlusion mask. Again, both the L1 norm and the total variation will be applied to regulate the extent of the deformations and the smoothness of the grid. As for the occlusion mask formulation, a downsampled version of the deformation grid will be considered for the optimization (since it is assumed that the optimization of fewer parameters can potentially lead to more robust results).


\chapter{Material and Experiments}

\section{Datasets} 
\label{sec:datasets}
Two different databases were considered for the evaluation of the proposed network architectures and visualization approaches. 
Classification performances were first assessed by performing an 8-fold cross validation for the weakly annotated data of the Computing in Cardiology (CinC) challenge 2017 \cite{clifford2017}. Moreover, the MIT-BIH Arrhythmia Database \cite{moody2001} was used to facilitate a visual evaluation of the attention approaches' localization abilities since it not only provides rhythm but also beat wise annotations and therefore more easily allowed to reason about attention outputs.

\subsection{The PhysioNet CinC Challenge 2017} 

The dataset of the PhysioNet Computing in Cardiology challenge 2017 \cite{clifford2017} consists of 12186 short ECG sequences that were recorded by AliveCor single-channel ECG devices. 
So far, a training set of 8528 records have been made available while the test set of 3658 records still remains private. In order to obtain scores for the hidden test set, classification models need to be submitted to the Physionet challenge community. The records of the CinC dataset have an average length of about 30 seconds (ranging from 9 - 61 seconds) and are sampled at 300 Hz. 
Experts manually classified each complete sequence into Normal rhythm (N), AF rhythm (AF), Other rhythm (O) and Noisy entries ($\sim$) and did refine their annotation several times during the official phase of the challenge. Since only global annotations are available for the network training, this work studies a weakly supervised learning task.

\subsection{The MIT-BIH Arrhythmia Database}

The MIT-BIH arrhythmia database \cite{moody2001} was provided by the Boston's Beth Israel Hospital and contains 48 two-channel records of 30 minutes duration. The records were obtained from 47 patients and are sampled at 360 Hz. The annotation was performed by at least two experts, which were assigning one from 15 different heartbeat types for each beat and furthermore provided annotations of signal quality changes, rhythm changes and the corresponding rhythm class. 

\section{Evaluation}
Since we had no access to the hidden test set, an 8-fold cross validation was performed on the training set to evaluate and compare the performances of studied models. Each validation fold consisted of 1066 records, resulting in a remaining training set of 7462 records for each of the eight evaluation runs. The class distribution of each fold was approximately 60\% Normal, 8\% AF,  29\% Other, and 3\% Noisy records. The performance was evaluated utilizing the overall F1 score, where the class F1 scores were computed for the classes $ c $ $\in \{AF, N, O, \sim\}$ using the following formula:
\begin{equation}
F1_{c} = \frac{2\times TP_{c}}{P_{c}+p_{c}}.
\end{equation}
Here, $ TP_{c} $ denotes the count of true positives, $P_{c}$ the count of positives, and $p_{c}$ the count of predicted positive records for the given class $c$. In accordance to the challenge guidelines, class F1 scores were then averaged over all classes except for class Noisy (which was severely underrepresented in the overall database):
\begin{equation}
F1 = \frac{F1_{AF}+F1_{N}+F1_{O}}{3}.
\end{equation}
Finally, the global F1 score was defined as the average F1 score over all folds.

\section{Network Architectures and Parameterisations}

This section gives an overview of studied CNN and ConvLSTM architectures. Notations in bold letters will thereby introduce the abbreviation of architectures that will be referred to in the next section. To handle varying record lengths in the CinC database and to facilitate the implementation of batch-processing, all input records were zero-padded to a length of 61 seconds. During training, the networks minimized the cross-entropy loss using stochastic gradient descent with Adam optimizer and batch sizes of 16. If not stated differently, the training was performed over $ 50 $ epochs using an initial learning rate of $ 0.001 $ that was decreased with a factor of $ 0.95 $ in each epoch. 

\subsection{Basic Convolutional Neural Network Modules} 

To ensure a fair performance comparison between global pooling and LSTM feature aggregation strategies, the following basic CNN networks were considered:

\paragraph{4 layer CNN module}
Throughout the experiments of this work, the application of shallow CNN architectures led to inferior results than the application of deeper networks. Nevertheless, a four layer CNN network (with 16, 32, 64, and 128 channels, each applying a kernel of size 21) was examined in order to investigate the influence of too weakly abstracted features on the performance of LSTM aggregation modules. Given a comparably large output size of $4535 \times 128$, 
the inference of global rhythm predictions appeared especially difficult for global pooling setups. The module in total contained 227366 trainable parameters. As in all other basic modules, each convolutional layer was followed by a batch normalization and a ReLU activation. 

\begin{figure}
\centering
\includegraphics[width=1.0\linewidth]{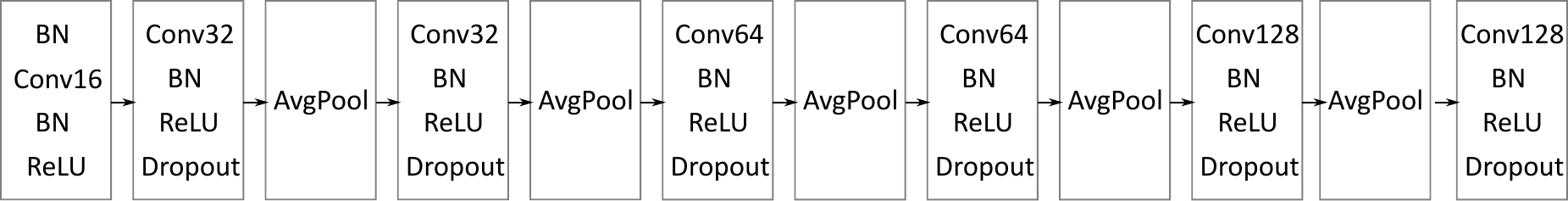}
\caption{Basic CNN network consisting of seven convolutional layers using batch normalization (BN) and intermediate average pooling layers (AvgPool).}
\label{fig:shallowBasisCNN}
\end{figure}

\paragraph{7 layer CNN module}
The second studied CNN architecture is illustrated in Fig.~\ref{fig:shallowBasisCNN}. It consisted of seven convolutional layers (with 16, 32, 32, 64, 64, 128, and 128 channels again using kernel sizes of 21) and overall 679622 trainable parameters. In order to speed up computations and to improve the generalization abilities of the network, average pooling (kernel size 2 and stride 2) was applied between all layers (except for the first and the last one). The dropout factor was increased layer after layer, giving the following dropout series (0, 0.2, 0.3, 0.4, 0.5, 0.5, 0). Considering padded records of size 18300 as input (corresponding to 61 seconds sampled at 300 Hz) the last layer feature map was of size $531 \times 128$ and thus assigned a 128 dimensional feature vector to each temporal output sample. 

\paragraph{15 layer CNN module}
The third basic CNN network was, with 15 convolutional layers in total, eight layers deeper than the previous module. Using kernel sizes of 21 and channel sizes of 16, 32, 32, 32, 32, 64, 64, 64, 64, 128, 128, 128, 128, 256, and 256, the module consisted of 3650566 trainable parameters. Applying average pooling after every second convolutional layer led to a network output size of $206 \times 256$.

\paragraph{15 layer CNN module with residual connections}
Inspired by a recent work of Rajpurkar et al.~\cite{rajpurkar2017}, another 15 layer CNN was studied using residual connections between blocks of CNN layers. In this module, the input of each CNN block was added to its output to facilitate a better gradient flow through the network. Since the convolutional layers were used without applying any padding, the feature maps that bypassed these convolutions had to be cropped to match the given output size. 

\paragraph{17 layer CNN module}
Given that the attention gated CNN approach required an even deeper architecture to successfully extract high level features at intermediate layers, a further 17 layer CNN was proposed using channel sizes of 16, 32, 32, 64, 64, 64, 64, 128, 128, 128, 128, 256, 256, 256, 256, 512, and 512. Again, batch normalization, ReLU activations and increasing dropout series were applied and average pooling was inserted after every second CNN layer. The last layer output finally consisted of 63 samples for the padded sequences and only 9 samples for the shortest 9 second records (when padded zero samples were removed after the convolution). 

\subsection{Global Pooling Setup} 
As introduced in Sec.~\ref{sec:methods:cnn}, global pooling layers can be stacked after the last CNN layer in order to temporally aggregate feature vectors for a subsequent classification layer (see Fig.~\ref{fig:cnn_net}). In this work, two different pooling variants were compared, namely global average pooling (\textbf{CNN+GAP}) and global max pooling (\textbf{CNN+GMP}).

\paragraph{Class activation maps}
The class activation maps that will be depicted in the next chapter, were computed as the weighted linear sum of the last CNN layer's upsampled feature maps and the weights that connect the maps to the neuron with the highest output activation (see Eq.~\ref{eq:cam}). Still, especially for cases of high prediction uncertainty, it can be helpful to also visualize CAMs for the remaining output neurons (resulting in attention maps showing evidences for the other classes).

The resolution of the class activation maps is mainly determined by the depth of the CNN network. That means feature maps of deep architectures are becoming increasingly coarse. Nevertheless to enable a visualization of `high resolution' attention maps, it was also experimented with incorporating additional global pooling layers for earlier feature maps (which show activations for lower-level features but are of higher resolution than the last layer feature map). For this purpose, the classification was trained on both the GMP vector of the last and the one of an additional intermediate layer of our choice (\textbf{15 layer CNN+GMP, mean vote}). 

\subsection{Gated Attention Setup}
As mentioned before, attention gated CNNs aim at extracting high level features at one or more intermediate feature scales. For this reason, a 17 layer deep CNN was studied that applied an attention gating at the 13th convolutional layer. The 13th layer was chosen specifically, since it provided outputs of size $492 \times 256$, which were of similar resolution than the output features of the 7 layer CNN module. 
As introduced by Schlemper et al.~\cite{schlemper2018}, attention weights were computed as similarity scores between the feature vectors of the intermediate and the last CNN layer (see Sec.~\ref{sec:methods:attentiongates}). Obtaining one score for each location, the feature maps were first multiplied (gated) in a channel-wise fashion with the attention weights matrix and finally temporally pooled to a vector of size $1 \times 256$. In order to aggregate the resulting vector with the one of the global pooling path (whose output was of size $1 \times 512$), two independent classification layers were trained. Subsequently, a mean vote strategy was applied to combine the scores of both paths.

\subsection{Convolutional Long Short-Term Memory Network Setup}

We experimented with three different ways of presenting the input data to the networks. First, extracting non-overlapping subwindows of 0.25 or 1 seconds. Second, using (overlapping) heart beat windows centered at detected R-peaks. And third, to present the ECG record at full length which was found to result in best performances (see Fig.~\ref{fig:cnn_lstm} for an illustration of the overall setup). 

\paragraph{Stacking CNN and LSTM modules}
As discussed in Sec.~\ref{sec:methods:cnnlstm}, each temporal CNN output sample $t$ is associated with one feature vector $x_{t}$ (whose dimensionality corresponds to the number of feature channels). An LSTM module that is stacked on a CNN module subsequently processes the feature sequence in order to find a hidden state representation that can be used for the classification. 

The `onedirectional' \textbf{LSTM}s of our experiments, performed many-to-one predictions by only passing the last time step hidden state to the classification layer. In further experiments, standard LSTMs were compared with bidirectional LSTMs and GRUs. Contrary to onedirectional LSTMs, \textbf{bidirectional LSTM}s concatenated the `last' hidden states of both directions (the backward module processed the input in reversed order) before passing the combined vector to the classification layer (see Fig.~\ref{fig:bidirectional_lstm}). An attempt to alternatively extract the hidden states of both directions at the central time step (to potentially solve memory problems for long sequences) is denoted as \textbf{CNN+bidirectional LSTM, center}. 
In a last setup (\textbf{CNN+bidirectional LSTM+pooling}), an additional connection between the CNN module and the classification layer was inserted by concatenating the last hidden state with a global pooling vector of the last CNN layer. Throughout the experiments it was found that a pretraining of CNN parameters using GMP and a classification layer potentially had a positive effect on the ConvLSTM performances (denoted as \textbf{pretrained} in the setup names). In those experiments, pretraining was performed over 50 epochs and the initial learning rate of the CNN parameters was reduced to $1e-4$ during the combined training with the LSTM parameters (which used an initial learning rate of $1e-3$).

\paragraph{Plotting class decisions over time} 
The many-to-one ConvLSTMs of this work performed one global class prediction after the whole record had been processed. Still, it can be helpful to examine the decision making process over time by also passing intermediate hidden states $h_{t}$ to the classification layer. By doing so, each time step $t \in {1,...,T}$ was assigned with softmax scores that could be plotted as the intermediate (attention-like) class confidences for a given input window (since the CNN output was of lower resolution that the original input). The `class decision over time' plots finally show the intermediate class decision corresponding to the maximum softmax score for each time step. 

\paragraph{Computation of shift perturbation masks} 
By minimizing the objective function of Eq.~\ref{eq:occlusionmask}, we aimed at finding input perturbations that were switching the network decision from class AF or Other to Normal. Beats of most salient rhythm irregularities could then be identified as samples showing the highest shift values in the perturbation mask. 
In order to ease the optimization problem, a downsampled version of the deformation grid was optimized that was computing shifts for only 1\%  
of the actual input samples (corresponding to a downsampling factor of 100). The downsampled grid was initialized with random shift values sampled from a normal distribution with zero mean and a standard deviation of $0.001$. Generally, the grid held values in the range $[-1,1]$ with sample locations being normalized by the lengths of the input (-1 corresponding to the very left input sample and 1 the very right one). Linear interpolation was used to sample the input pixels and to derive a gradient for optimization. Optimization again was performed with stochastic gradient descent and Adam optimizer and aimed at the computation of an optimal perturbation mask that flipped the class decision of the network. This time, a smaller learning rate of $0.0001$ and an optimizing over 500 epochs were considered. The hyperparameters $\lambda_{1}$ (L1 coefficient) and $\lambda_{1}$ (TV coefficient) influenced the extent of the shifts and the smoothness of the final perturbation grid result. Without any regularization the perturbations did not produce helpful visualizations and flipped the label to the class Noisy in most cases.

\paragraph{Visualization of hidden states and gates}
Karpathy \cite{karpathy2015} and Chen \cite{chen2017} recently managed to gain some insights into the LSTM cell behavior by analyzing its states and input, forget, and output gates. However, Karpathy \cite{karpathy2015} found that for the prediction of next characters in creating programming code only 5\% of the neurons performed meaningful operations (like tracking the position in line, turning on in quotes, activating inside if statements, etc.) and that many neuron states remained hard to interpret. For this reason, the internal gate and state values for a very simple 2 hidden unit LSTM will be analyzed. Values for each time step gate and state values were thereby computed according to the equations \eqref{eq:lstmfirst} to \eqref{eq:lstmlast}.

\section{Implementation and Computation Times}

All models were implemented in Python using the Pytorch library \cite{paszke2017} and each experiment was run on a single GPU (NVIDIA GeForce GTX 1080 Ti, 11GB RAM). Computation times depend on the network setups and particularly the length of the feature sequence that was processed by the LSTM module (and that the error needed to be backpropagated through). While most of the models (both CNNs with global pooling and ConvLSTMs) took between 1 to 3 hours to train, 
much longer training times of about 4 to 5 hours were observed for the ConvLSTM that extracted features using a shallow 4 layer CNN setup and therefore processed longer feature sequences. Certainly, training times were also influenced by the model complexities. The number of trainable parameters were 679622 for the 7 layer CNN, 3650566 for the deeper 15 layer CNN, 3732230 for the ConvLSTM with 15 CNN layers and 64 hidden units, and finally 3765510 for the ConvLSTM consisting of 15 CNN layers and 2 LSTM layers with 64 hidden units.

\chapter{Results}

This section reports the cross validation performances of the previously introduced network architectures and illustrates various attention maps for a selection of models. The presented tables give an overview of class specific and total F1 scores that have been averaged over all folds of the cross validation. As introduced in the last chapter, the overall F1 score does not include the class Noisy.

\section{Global Pooling Performances} 
\begin{table}
\centering
\caption{8-fold cross validation of global pooling setups}
\label{tab:evaluation:cnn}
\resizebox{\textwidth}{!}{%
\begin{threeparttable}[b]
\begin{tabular}{@{}rrrrrr@{}}
\toprule
Architecture & $F1_{AF}$ & $F1_{N}$  & $F1_{O}$  & $F1_{\sim}$  & $F1_{total}$  \\ \midrule
4 layer CNN+GMP & 0.61 ($ \pm $ 0.06) & 0.86 ($ \pm $ 0.03) & 0.59 ($ \pm $ 0.03) & 0.50 ($ \pm $ 0.09) &  0.68 ($ \pm $ 0.04) \\
7 layer CNN+GAP & 0.79 ($ \pm $ 0.05) & 0.91 ($ \pm $ 0.01) & 0.75 ($ \pm $ 0.03) & 0.60 ($ \pm $ 0.07) &  0.82 ($ \pm $ 0.02) \\
7 layer CNN+GMP  & 0.78 ($ \pm $ 0.03) & 0.92 ($ \pm $ 0.01) & 0.76 ($ \pm $ 0.02) & 0.56 ($ \pm $ 0.08) &  0.82 ($ \pm $ 0.01) \\
15 layer CNN+GMP  & 0.82 ($ \pm $ 0.02) & 0.92 ($ \pm $ 0.01) & 0.79 ($ \pm $ 0.02) & 0.63 ($ \pm $ 0.07) &  \textbf{0.84} ($ \pm $ 0.01) \\
15 layer, residual CNN+GMP  & 0.80 ($ \pm $ 0.03) & 0.91 ($ \pm $ 0.01) & 0.77 ($ \pm $ 0.02) & 0.63 ($ \pm $ 0.05) &  0.83 ($ \pm $ 0.02) \\
15 layer CNN+GMP, mean vote  & 0.81 ($ \pm $ 0.02) & 0.91 ($ \pm $ 0.01) & 0.78 ($ \pm $ 0.02) & 0.54 ($ \pm $ 0.11) &  0.83 ($ \pm $ 0.02) \\
17 layer CNN+GMP & 0.81 ($ \pm $ 0.04) & 0.92 ($ \pm $ 0.01) & 0.78 ($ \pm $ 0.02) & 0.65 ($ \pm $ 0.04) &  0.83 ($ \pm $ 0.02) \\
\bottomrule
\end{tabular}
\end{threeparttable}
}
\end{table}

In first experiments, global rhythm classifications were performed by global pooling CNNs. 
Table \ref{tab:evaluation:cnn} gives an overview of F1 scores that were obtained for CNN modules of varying depths. Despite a prior assumption that GMP would outperform GAP due to a superior capturing of episodic abnormalities, both pooling variants obtained an equal average F1 score of $0.82$ when studying a 7 layer CNN module. Performances were further improved to a score of $0.84$ when using a deeper network setup with 15 layers and 256 output channels (\textbf{15 layer CNN+GMP}). The insertion of residual connections (\textbf{15 layer, residual CNN+GMP}), however, even slightly dropped the F1 score to $0.83$. Furthermore, it can be observed that the performances with $F1=0.68$ were significantly inferior for the 4 layer CNN setup (\textbf{4 layer CNN+GMP}) where the temporal aggregation of features had to be performed over long sequences of $4535$ output samples.

\begin{figure}
\centering
  \begin{subfigure}[b]{1.0\textwidth}
  	\includegraphics[width=1.0\linewidth]{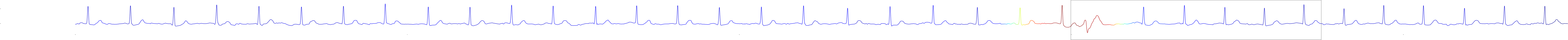}
  \end{subfigure}
  \begin{subfigure}[b]{1.0\textwidth}
	\includegraphics[width=1.0\linewidth]{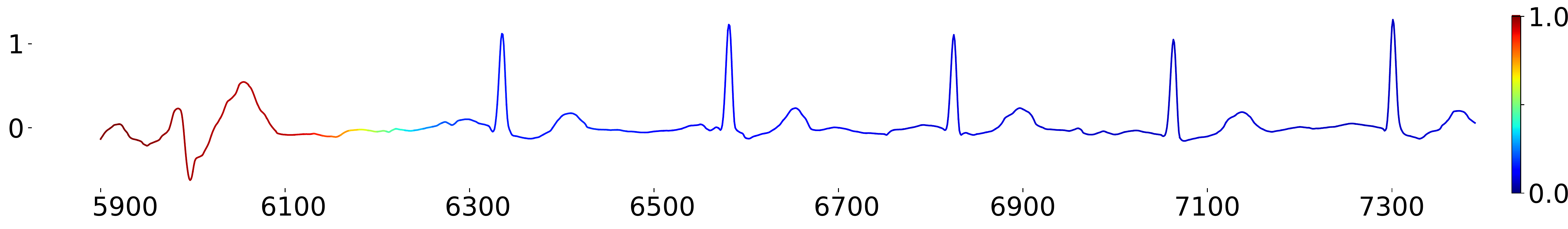}
  \end{subfigure}
  \caption{CNN+GAP CAM for the correctly classified Other rhythm record A06020. Full record visualization (top) and 5 second excerpt (bottom) of an attention map that successfully highlights an abnormal beat. As indicated by the color bar, activations are color coded from high to low as red to blue.}
  \label{fig:cnn_avgpool_example3}	
\end{figure}

\begin{figure}
\centering
  \begin{subfigure}[b]{1.0\textwidth} 
  	\includegraphics[width=1.0\linewidth] {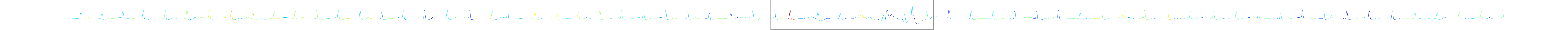}
  \end{subfigure}
  \begin{subfigure}[b]{1.0\textwidth}
  	\includegraphics[width=1.0\linewidth, trim={0 14cm 0 0}, clip]{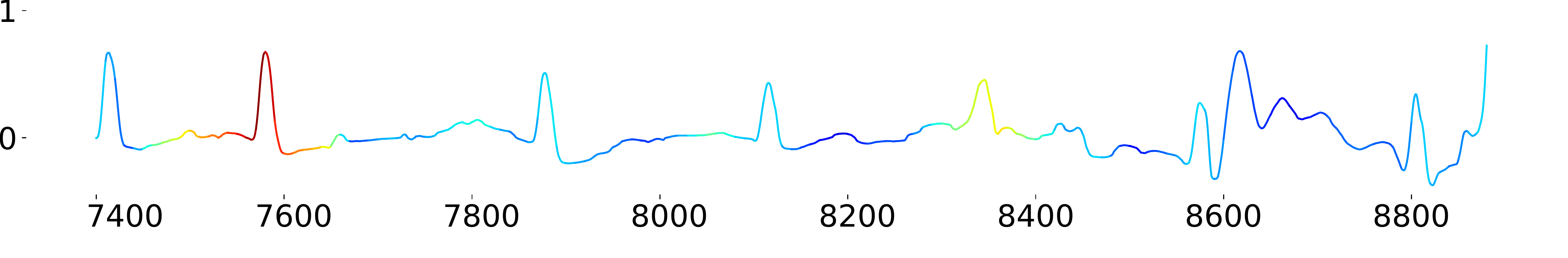}
  \end{subfigure}
   \caption{CNN+GMP CAM for the correctly classified Other rhythm record A05831. The attention map particularly focuses on one premature beat while assigning the lowest activations to a noisy section at the center of the record (visible in the full record visualization at the top).}
   \label{fig:cnn_maxpool_example1}
\end{figure}

\subsection{Class Activation Map Visualizations}

\begin{figure}[H]
\centering
  \begin{subfigure}[b]{1.0\textwidth}
  	\includegraphics[width=1.0\linewidth]{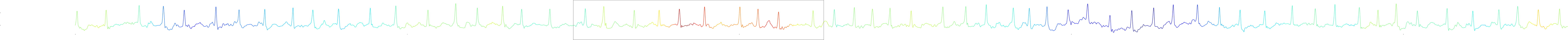}
  \end{subfigure}
  \begin{subfigure}[b]{1.0\textwidth}
	\includegraphics[width=1.0\linewidth, trim={0 4cm 0 0},]{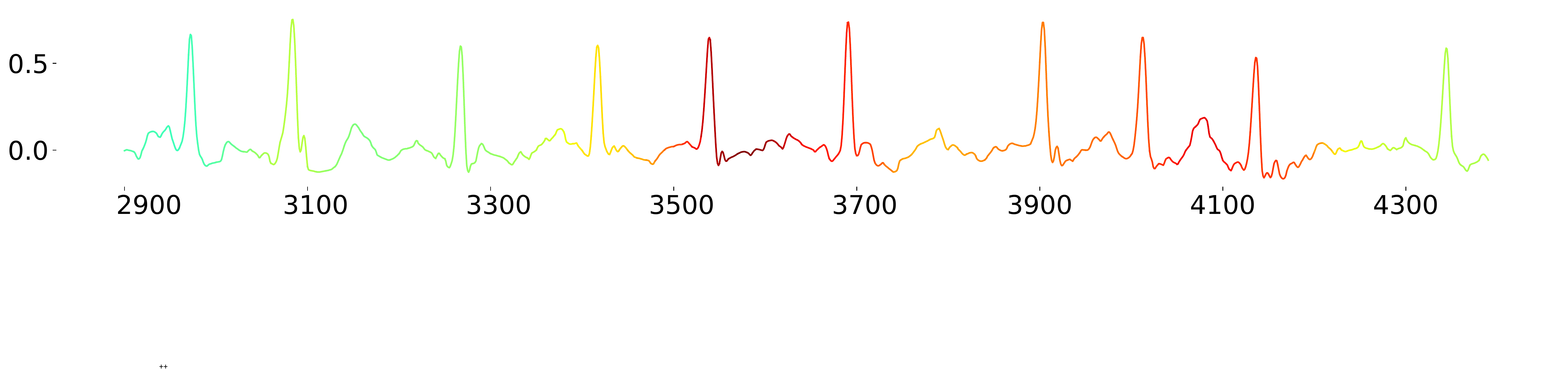}
  \end{subfigure}
  	\caption{CNN+GAP CAM for the correctly classified AF rhythm record A00225. The attention map highlights some (but not all) beats with irregular RR interval.} 
  \label{fig:cnn_avgpool_example1}
\end{figure}

\begin{figure}[H]
\centering
  \begin{subfigure}[b]{1.0\textwidth}
  	\includegraphics[width=1.0\linewidth]{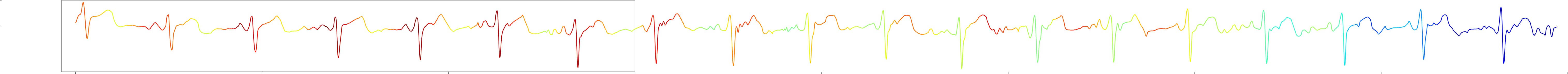}
  \end{subfigure}
  \begin{subfigure}[b]{1.0\textwidth}
	\includegraphics[width=1.0\linewidth]{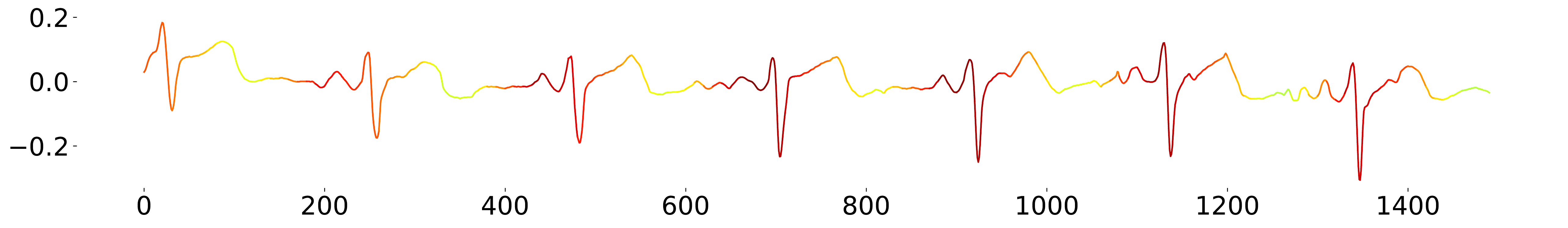}
   \end{subfigure}
  	\caption{CNN+GAP CAM for the correctly classified Normal rhythm record A00464. Apparently, the network focused on several beats of similar appearance equally.}
  	\label{fig:cnn_avgpool_example2}
\end{figure}

\begin{figure}[H]
\centering
  \begin{subfigure}[b]{1.0\textwidth}
  	\includegraphics[width=1.0\linewidth]{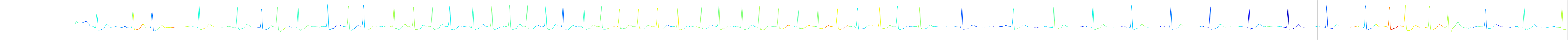}
  \end{subfigure}
  \begin{subfigure}[b]{1.0\textwidth}
  	\includegraphics[width=1.0\linewidth]{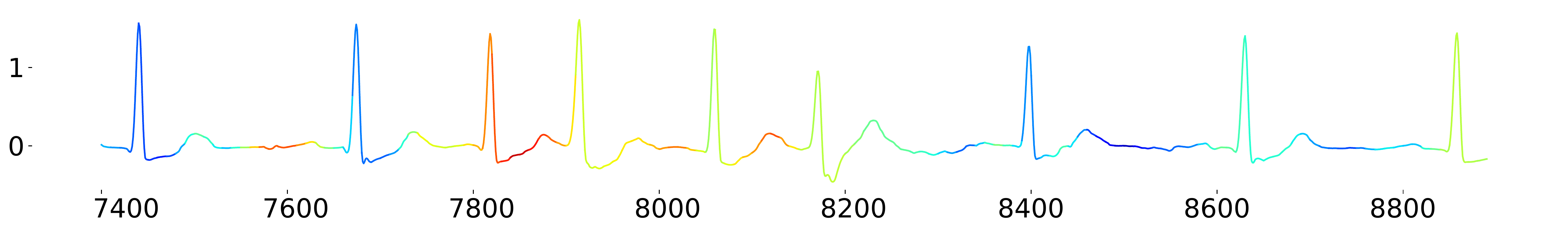}
  \end{subfigure}
   \caption{CNN+GMP CAM for the AF rhythm record A01718 that was incorrectly predicted as Other rhythm.} 
   \label{fig:cnn_maxpool_example2}
\end{figure}

\begin{figure}[H]
\centering
  \begin{subfigure}[b]{1.0\textwidth}
  	\includegraphics[width=1.0\linewidth]{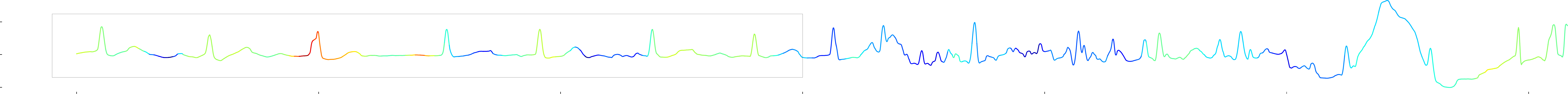}
  \end{subfigure}
  \begin{subfigure}[b]{1.0\textwidth}
  	\includegraphics[width=1.0\linewidth]{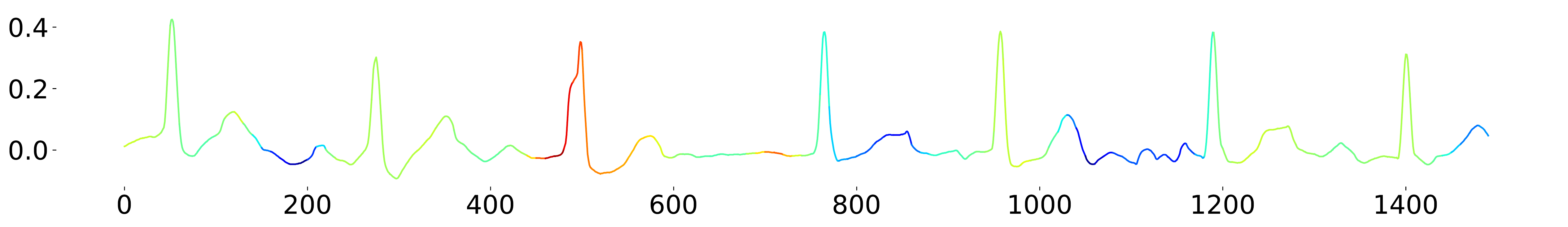}
  \end{subfigure}
  \caption{CNN+GMP CAM for the misclassified Noisy rhythm record A08043. Even though large parts of the record were of noisy appearance, the clear detection of AF specific rhythm irregularities caused an AF class prediction. This case illustrates a weakness of the global max pooling operation, which is not able to capture information about the duration of feature occurrences.}  
   \label{fig:cnn_maxpool_example3}	
\end{figure}

The influence of the pooling variant choice can be further examined by studying the class activation maps that were computed for a network with 7 convolutional layers (see Fig.~\ref{fig:cnn_avgpool_example1} to Fig.~\ref{fig:cnn_maxpool_example3}). Each depicted attention map represents the CAM for the output neuron with the highest softmax score and therefore highlights input patterns that were considered to be meaningful evidences for the predicted class.

The first record example of Fig.~\ref{fig:cnn_avgpool_example3} shows a class activation map that resulted from an GAP architecture. The visualization suggests that the network was confident about the class prediction Other and that one particular abnormal beat was highlighted in the rhythm. Despite the episodic character of the pathological event, the corresponding section of high CAM activations (indicated by the red color coding) dominated all remaining temporal samples in the averaging process and provoked the correct class prediction Other on the global level. 

While Other rhythm examples often exhibited strong activations for single prominent record sections, interpretations were generally less intuitive for examples of classes AF and Normal. In those cases, where class specific patterns occurred repeatedly, CAMs either assigned high activations for many subsequent beats (like in Fig.~\ref{fig:cnn_avgpool_example2}) or only focused on single pattern occurrences. The figure \ref{fig:cnn_avgpool_example1}, for instance, shows a CAM, where an irregular RR interval feature was highlighted at the central part but not at the beginning of the record. This observation indicates that final rhythm predictions can base solely on the detection of single key features of strong evidence and are rarely taking all episodes of pathologies into account. An additional, more encouraging observation was the assignment of very low activation for class irrelevant patterns (colored in dark blue), as it can be seen in the noisy section of Fig.~\ref{fig:cnn_avgpool_example1}.

\begin{figure}[H]
\centering
  \begin{subfigure}[b]{1.0\textwidth}
  	\includegraphics[width=1.0\linewidth]{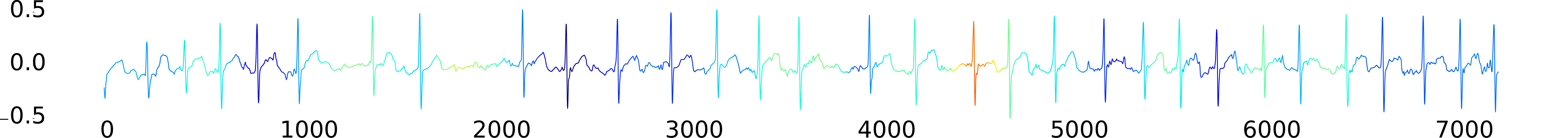}
  \end{subfigure}
  \begin{subfigure}[b]{1.0\textwidth}
  	\includegraphics[width=1.0\linewidth]{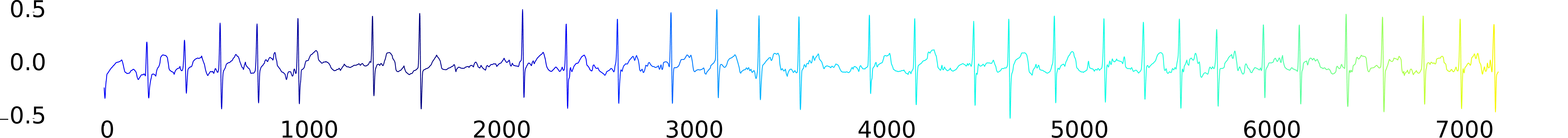}
  \end{subfigure}
  	\caption{15 layer CNN+GMP CAMs (obtained by a concatenation strategy) for the correctly classified AF record A01718. The visualizations show the CAMs that were computed for the intermediate (top) and the last layer (bottom).}
  	\label{fig:concat_pooled1}
\end{figure}
\begin{figure}
\centering
  \begin{subfigure}[b]{1.0\textwidth}
  	\includegraphics[width=0.98\linewidth, trim={0 0 3cm 0}, clip]{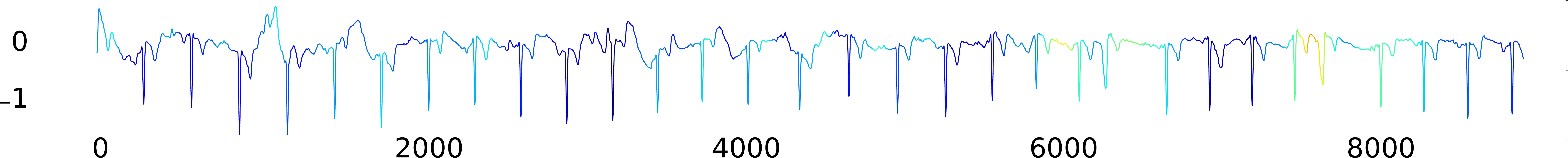}
  \end{subfigure}
  \begin{subfigure}[b]{1.0\textwidth}
  	\includegraphics[width=0.98\linewidth, trim={0 0 3cm 0}, clip]{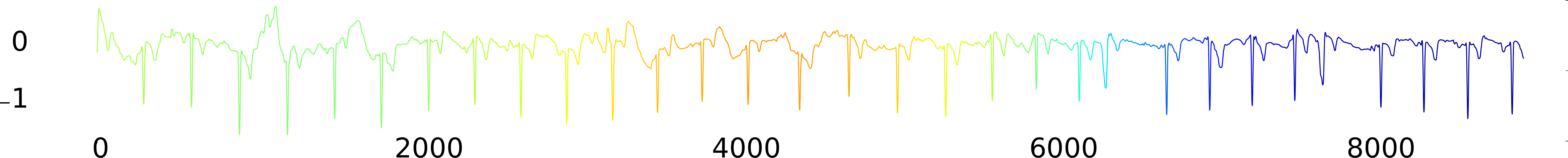}
  \end{subfigure}
  	\caption{15 layer CNN+GMP CAMs (obtained by a mean vote strategy) for the correctly classified Other record A03516. Again, the visualizations show the CAMs that were computed for the intermediate (top) and the last layer (bottom).}
  	\label{fig:meanvote_pooled1}
\end{figure}
In cases of misclassifications, class activation maps can also help to closer examine record patterns that caused wrong predictions. Figure \ref{fig:cnn_maxpool_example3}, for instance, shows a class activation map of a misclassified record of ground truth class Noisy that clearly suggests that the classification as AF was triggered by the detection of missing P waves and irregular rhythms. Deciding at which extent and amount of noise a record was to be classified as Noisy appeared to be a non trivial and potentially ambiguous task and was discussed exhaustingly during the challenge (leading to several refinements of the expert annotations).

As discussed in the experimental setup description, another attempt to improve the classification performance was to explicitly consider global as well as more local information when performing the classification (\textbf{15 layer CNN+GMP, mean vote}). Nevertheless, neither a concatenation of both vectors nor the fitting of two independent classification layers (whose outputs were combined by averaging) showed any performance improvements and even dropped the F1 score to $ 0.83 $. One remaining benefit of such intermediate CAM computations was the ability to visualize informative attention maps for deep networks. Two examples of intermediate layer CAMs are depicted in Fig.~\ref{fig:concat_pooled1} and Fig.~\ref{fig:meanvote_pooled1} (showing correctly classified records of classes AF and Other, respectively). While the CAM of the intermediate layer apparently highlights single, suspicious beats, the coarse CAM (computed for the last layer) hardly gives any interpretable insights regarding the class decision of the network. 

\section{Gated Attention Network Performances}

\begin{figure}
\centering
  \begin{subfigure}[b]{1.0\textwidth}
  	\includegraphics[width=1.0\linewidth]{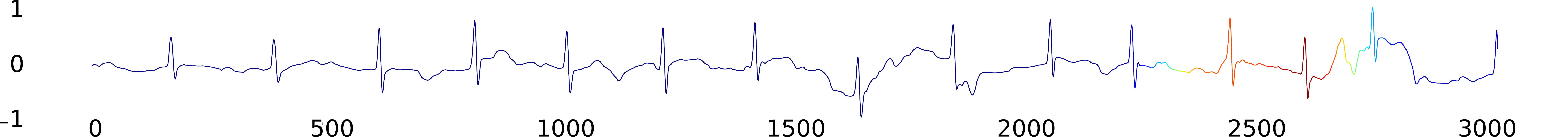}
  \end{subfigure}
  \begin{subfigure}[b]{1.0\textwidth}
  	\includegraphics[width=1.0\linewidth]{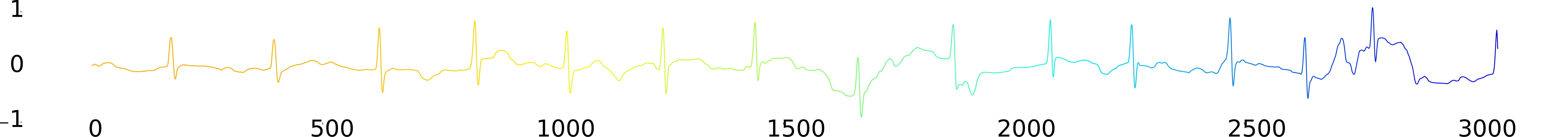}
	\caption{} 
	\label{fig:attention_gate_a}
  \end{subfigure}
  \begin{subfigure}[b]{1.0\textwidth}
  	\includegraphics[width=1.0\linewidth]{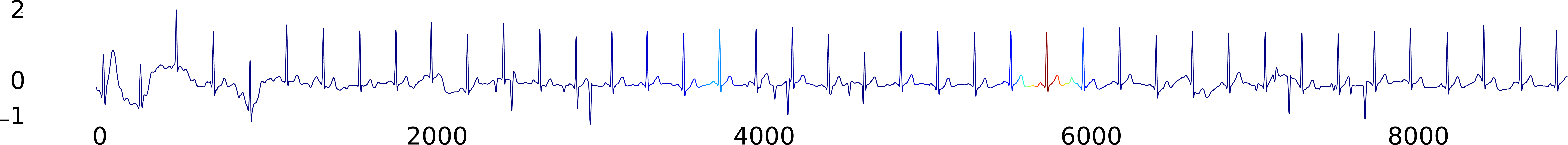}
  \end{subfigure}
  \begin{subfigure}[b]{1.0\textwidth}
  	\includegraphics[width=1.0\linewidth]{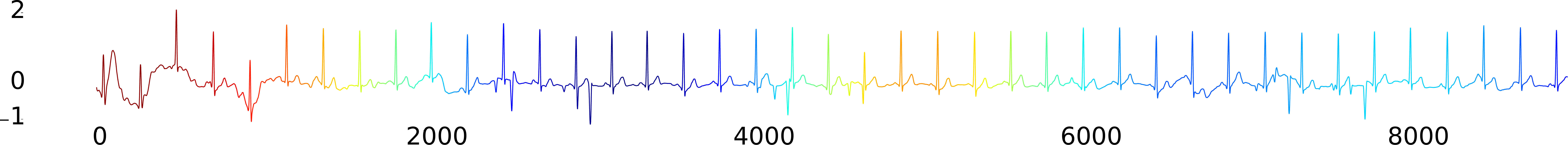}
   \caption{} 
  \label{fig:attention_gate_b}
  \end{subfigure}
    \begin{subfigure}[b]{1.0\textwidth}
  	\includegraphics[width=1.0\linewidth]{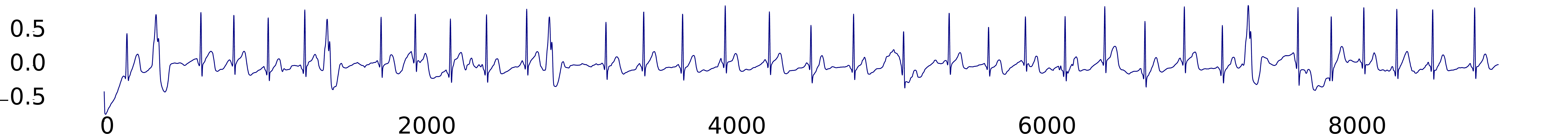}
  \end{subfigure}
  \begin{subfigure}[b]{1.0\textwidth}
  	\includegraphics[width=1.0\linewidth]{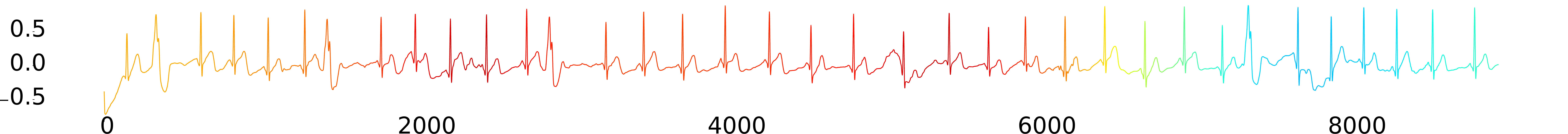}
   \caption{} 
   \label{fig:attention_gate_c}
  \end{subfigure}
  \caption{Gated attention maps (top) and CAMs (bottom) for examples of classes AF (a), Normal (b), and Other (c) that were all correctly classified by the attention network. While the gated attention maps were extracted at the 13th layer, the CAMs were computed for the output neuron that corresponded to the highest softmax score. The network was trained with a mean vote aggregation strategy to combine the scores of two independently trained classification layers. While the attention gated feature maps for both AF and Normal example focus on one to two particular beats, the feature map for the class Other example does not show any high activations throughout the entire record.} 
  \label{fig:meanvote_conv7_conv9}
\end{figure}

The results in Table \ref{tab:evaluation:cnn_attentiongates} indicate that the extension of a global pooling CNN with an additional attention path could slightly improve the performance from $0.83$ to $0.84$ when using two independent classification layers for both paths. First attempts to train a single classification layer for a concatenated output of both paths resulted in `empty' attention maps for the intermediate layer and consequently a skipping of the attention gating path. Figure \ref{fig:meanvote_conv7_conv9} shows some examples of resulting intermediate gated attention maps and last layer CAMs that were computed using GMP and mean voting over two classifiers. While the attention map of the 13th layer highlights single beats, the CAM of the last layer was of much coarser appearance. As for the concatenation strategy setup, class Other record attention maps strikingly often did not show any high activations (indicating that local features were not considered for the classification). In fact, outputs of attention gated CNNs generally appeared less interpretable than the standard CAMs presented in the previous section. 

\begin{table}
\centering
\caption{8-fold cross validation of gated attention CNN setups}
\label{tab:evaluation:cnn_attentiongates}
\resizebox{\textwidth}{!}{%
\begin{threeparttable}[b]
\begin{tabular}{@{}rrrrrr@{}}
\toprule
Architecture & $F1_{AF}$ & $F1_{N}$  & $F1_{O}$  & $F1_{\sim}$  & $F1_{total}$  \\ \midrule
17 layer CNN+GMP & 0.81 ($ \pm $ 0.04) & 0.92 ($ \pm $ 0.01) & 0.78 ($ \pm $ 0.02) & 0.65 ($ \pm $ 0.04) &  0.83 ($ \pm $ 0.02) \\
17 layer CNN+AttentionGates, concat & 0.81 ($ \pm $ 0.05) & 0.92 ($ \pm $ 0.01) & 0.78 ($ \pm $ 0.02) & 0.66 ($ \pm $ 0.05) &  0.83 ($ \pm $ 0.02)  \\
17 layer CNN+AttentionGates, mean vote & 0.81 ($ \pm $ 0.04) & 0.92 ($ \pm $ 0.01) & 0.78 ($ \pm $ 0.02) & 0.63 ($ \pm $ 0.09) &  \textbf{0.84} ($ \pm $ 0.02)  \\
\bottomrule
\end{tabular}
\end{threeparttable}
}
\end{table}

\section{Convolutional Long Short-Term Memory Network Performances}

This section studies the influence of CNN and LSTM parameter choices when combining both modules to ConvLSTMs. An overview of all considered parameterizations (where \textbf{4 layer CNN}, \textbf{7 layer CNN}, \textbf{15 layer CNN}, and \textbf{15 layer, residual CNN} again denote the basic CNN networks) is given in Table \ref{tab:evaluation:lstm}.

\subsection{Sensitivity to Hyperparameters} 
Recapitulating all experiments, the network depth of the CNN module appeared to have a larger effect on the performance than the number of layers or hidden units concerning the LSTM module. While the shallowest setup with four convolutional layers and 64 hidden units reached an F1 score of $0.75$, the ConvLSTM consisting of 7 CNN layers and only 4 LSTM units already yielded a score of $0.82$. Increasing the dimension of the hidden state to 16 slightly improved the performance to $0.83$ and using multiple layers in combination with a pretraining of CNN parameters finally yielded a score of $0.84$.
Apparently, neither the bidirectional LSTM variant nor the application of a simplified GRU module brought any benefits compared to the 1 layer LSTM setup. The attempt to pass pooled CNN features as well as LSTM outputs to the classification layer (see setup \textbf{7 layer CNN+bidirectional LSTM+pooling}) even resulted in a slight performance drop which implies that the additional path of information had a negative impact on the training. 

The best ConvLSTM performance of $0.85$ was finally obtained with the setup \textbf{pretrained 15 layer CNN+1 layer LSTM, 64 hidden}. However, the application of a t-test confirmed that performances were not significantly better than other setups that reached a score of $0.84$ for the same basic CNN module. In conclusion, the positive effect of LSTMs on the classification accuracy is more pronounced for networks with fewer layers and the influence of the LSTM parameter choices decreased with growing depth.

 \begin{table}
 \centering
 \caption{8-fold cross validation of ConvLSTM setups}
 \label{tab:evaluation:lstm}
 \resizebox{\textwidth}{!}{%
 \begin{threeparttable}[b]
 \begin{tabular}{@{}rrrrrr@{}}
 \toprule
 Architecture & $F1_{AF}$ & $F1_{N}$  & $F1_{O}$  & $F1_{\sim}$  & $F1_{total}$  \\ \midrule
4 layer CNN+1 layer LSTM, 16 hidden  & 0.63 ($ \pm $ 0.04) & 0.86 ($ \pm $ 0.01)  &  0.59 ($ \pm $ 0.04)  & 0.44 ($ \pm $ 0.12) & 0.69 ($ \pm $ 0.03) \\
4 layer CNN+1 layer LSTM, 64 hidden  & 0.69 ($ \pm $ 0.04) & 0.88 ($ \pm $ 0.01)  &  0.68 ($ \pm $ 0.03)  & 0.56 ($ \pm $ 0.10) & 0.75 ($ \pm $ 0.02) \\
 \midrule
7 layer CNN+1 layer LSTM, 4 hidden  & 0.77 ($ \pm $ 0.04) & 0.91 ($ \pm $ 0.01)  &  0.77 ($ \pm $ 0.02)  & 0.56 ($ \pm $ 0.12) & 0.82 ($ \pm $ 0.02) \\
7 layer CNN+1 layer LSTM, 16 hidden & 0.80 ($ \pm $ 0.03) & 0.92 ($ \pm $ 0.01)  &  0.78 ($ \pm $ 0.02)  & 0.65 ($ \pm $ 0.06) & 0.83 ($ \pm $ 0.02) \\
7 layer CNN+1 layer LSTM, 64 hidden & 0.80 ($ \pm $ 0.03) & 0.92 ($ \pm $ 0.01)  &  0.78 ($ \pm $ 0.02)  & 0.64 ($ \pm $ 0.08) & 0.83 ($ \pm $ 0.02) \\
 pretrained 7 layer CNN+2 layer LSTM, 16 hidden & 0.81 ($ \pm $ 0.02) & 0.92 ($ \pm $ 0.01) & 0.78 ($ \pm $ 0.02)& 0.65 ($ \pm $ 0.06) & 0.84 ($ \pm $ 0.02)\\
 pretrained 7 layer CNN+3 layer LSTM, 32 hidden & 0.82 ($ \pm $ 0.02) & 0.92 ($ \pm $ 0.01) & 0.78 ($ \pm $ 0.02)& 0.65 ($ \pm $ 0.10) & 0.84 ($ \pm $ 0.01) \\
 7 layer CNN+bidirectional LSTM, 16 hidden & 0.81 ($ \pm $ 0.02) & 0.92 ($ \pm $ 0.01)  &  0.78 ($ \pm $ 0.03)  & 0.65 ($ \pm $ 0.08) & 0.83 ($ \pm $ 0.02) \\
7 layer CNN+bidirectional LSTM, 16 hidden, center & 0.80 ($ \pm $ 0.04) & 0.92 ($ \pm $ 0.01)  &  0.78 ($ \pm $ 0.02)  & 0.65 ($ \pm $ 0.07) & 0.83 ($ \pm $ 0.02) \\
 7 layer CNN+bidirectional GRU, 16 hidden & 0.81 ($ \pm $ 0.02) & 0.91 ($ \pm $ 0.01)  &  0.77 ($ \pm $ 0.02)  & 0.64 ($ \pm $ 0.06) & 0.83 ($ \pm $ 0.01) \\
7 layer CNN+bidirectional LSTM, 64 hidden+pooling & 0.80 ($ \pm $ 0.02) & 0.91 ($ \pm $ 0.01) & 0.76 ($ \pm $ 0.02) & 0.59 ($ \pm $ 0.07) &  0.82 ($ \pm $ 0.01) \\
 \midrule
  pretrained 15 layer CNN+1 layer LSTM, 2 hidden & 0.79 ($ \pm $ 0.04) & 0.91 ($ \pm $ 0.01) & 0.78 ($ \pm $ 0.02) & 0.52 ($ \pm $ 0.21) &  0.83 ($ \pm $ 0.02) \\ 
 pretrained 15 layer CNN+1 layer LSTM, 4 hidden & 0.82 ($ \pm $ 0.03) & 0.92 ($ \pm $ 0.01) & 0.78 ($ \pm $ 0.02) & 0.62 ($ \pm $ 0.06) &  0.84 ($ \pm $ 0.02) \\ 
 pretrained 15 layer CNN+1 layer LSTM, 64 hidden & 0.83 ($ \pm $ 0.03) & 0.92 ($ \pm $ 0.01) & 0.79 ($ \pm $ 0.02) & 0.64 ($ \pm $ 0.06) &  \textbf{0.85} ($ \pm $ 0.02) \\ 
 pretrained 15 layer CNN+2 layer LSTM, 64 hidden  & 0.82 ($ \pm $ 0.03) & 0.92 ($ \pm $ 0.01) & 0.79 ($ \pm $ 0.02) & 0.64 ($ \pm $ 0.09) &  0.84 ($ \pm $ 0.02) \\
 pretrained 15 layer residual CNN+2 layer LSTM, 64 hidden  & 0.80 ($ \pm $ 0.02) & 0.91 ($ \pm $ 0.01) & 0.78 ($ \pm $ 0.02) & 0.60 ($ \pm $ 0.06) &  0.83 ($ \pm $ 0.01) \\
 \bottomrule
 \end{tabular}
 \end{threeparttable}
 }
 \end{table}
 
\subsection{Plotting Class Decisions Over Time} 
\begin{figure}
	\centering
	\begin{subfigure}[b]{0.49\textwidth}
	\includegraphics[width=1.0\linewidth]{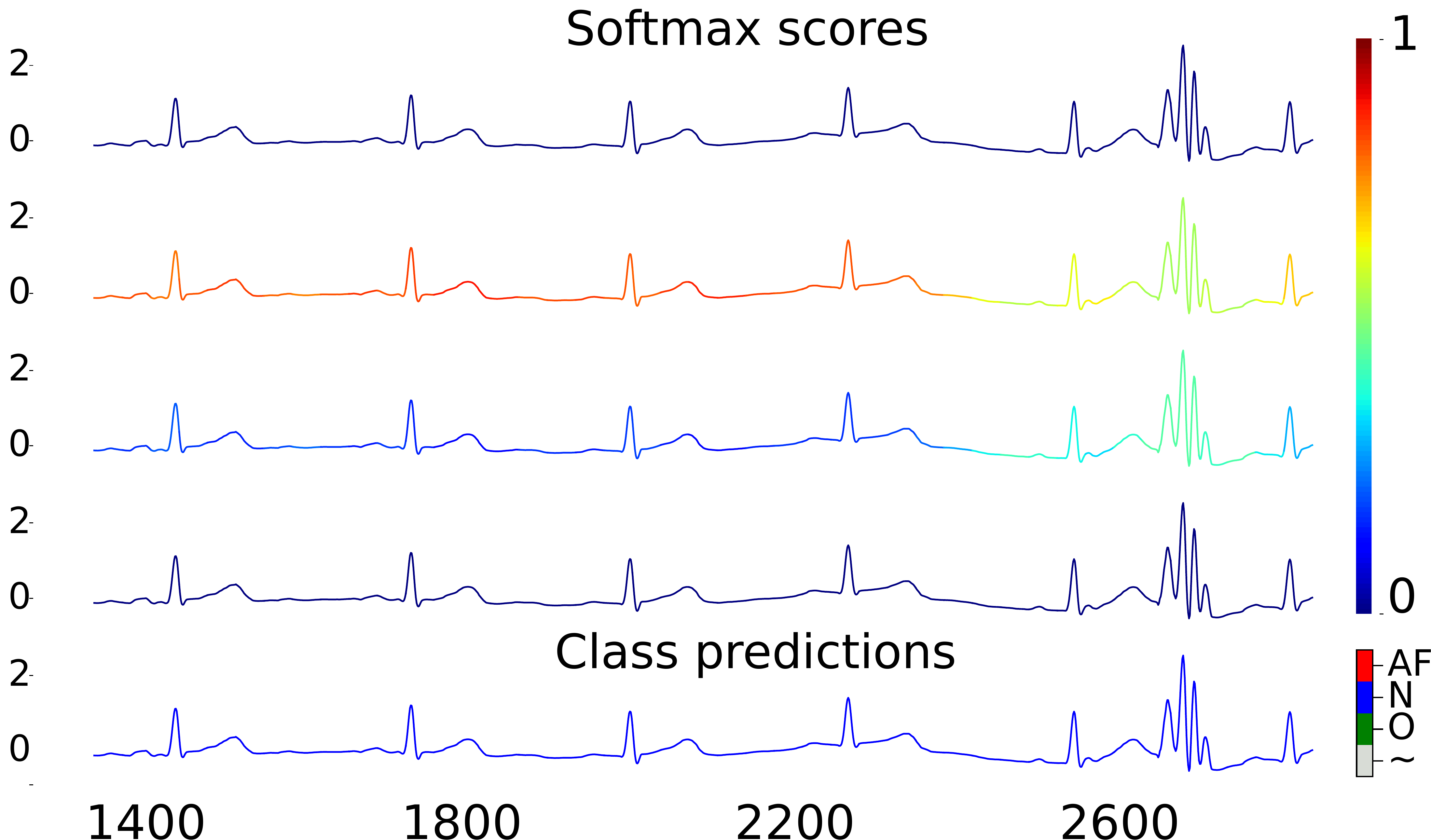}
	\caption{}
	\label{fig:lstm_timeplot0}
	\end{subfigure}
	\begin{subfigure}[b]{0.49\textwidth}
	\includegraphics[width=1.0\linewidth]{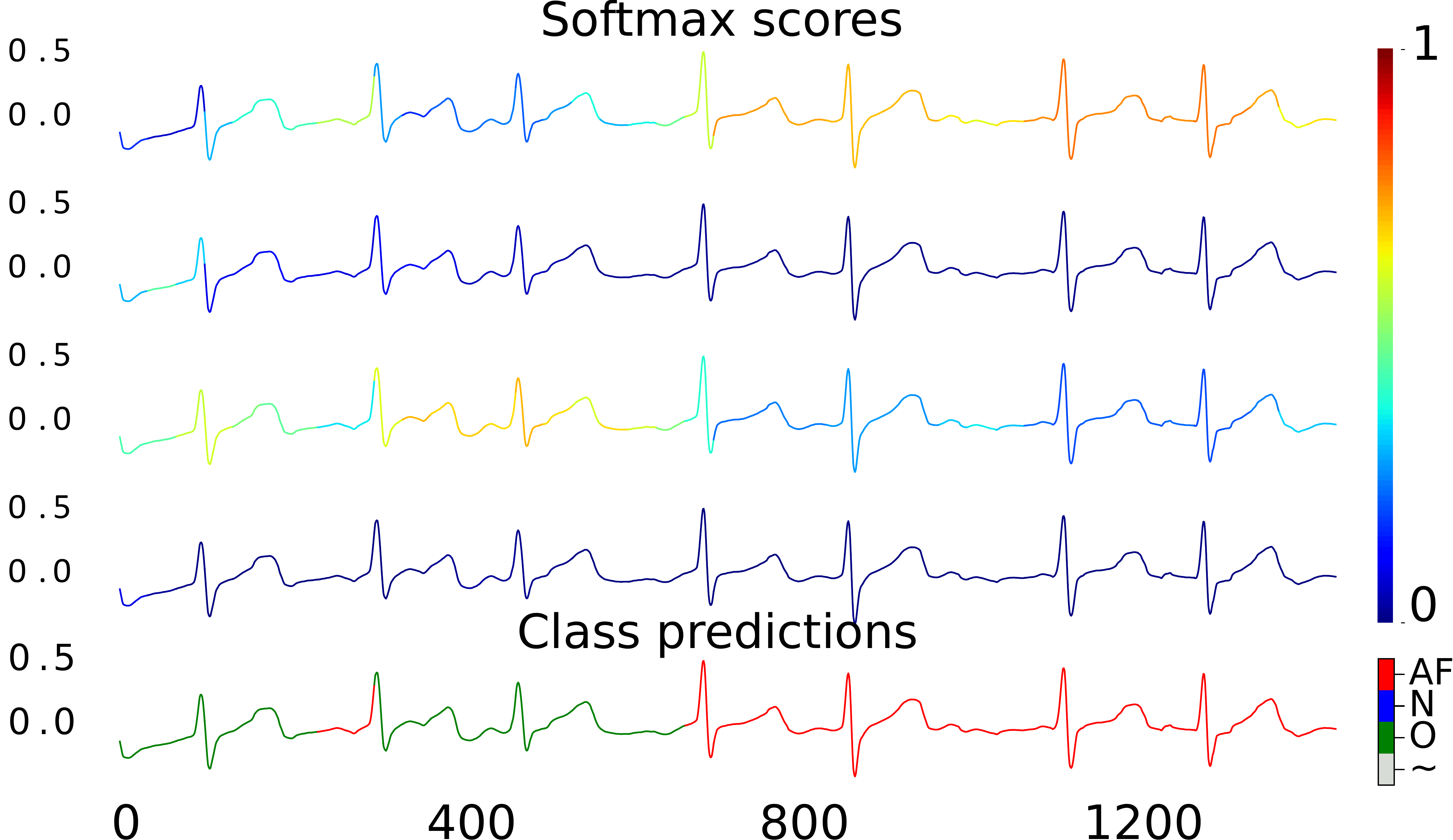}
	\caption{}
	\label{fig:lstm_timeplot1}
	\end{subfigure}
	\caption{5 second excerpt plot of intermediate class decisions for an example of class Normal (a) and class AF (b). While the plots 1-4 (from top to bottom) show the softmax scores changing over time for classes AF, N, O, and Noisy, plot 5 depicts the resulting class decisions (with classes being color coded according to the legend at the right).} 
	\label{fig:timeplot1}
\end{figure}

\begin{figure}
	\centering
	\begin{subfigure}[b]{0.49\textwidth}
	\includegraphics[width=1.0\linewidth]{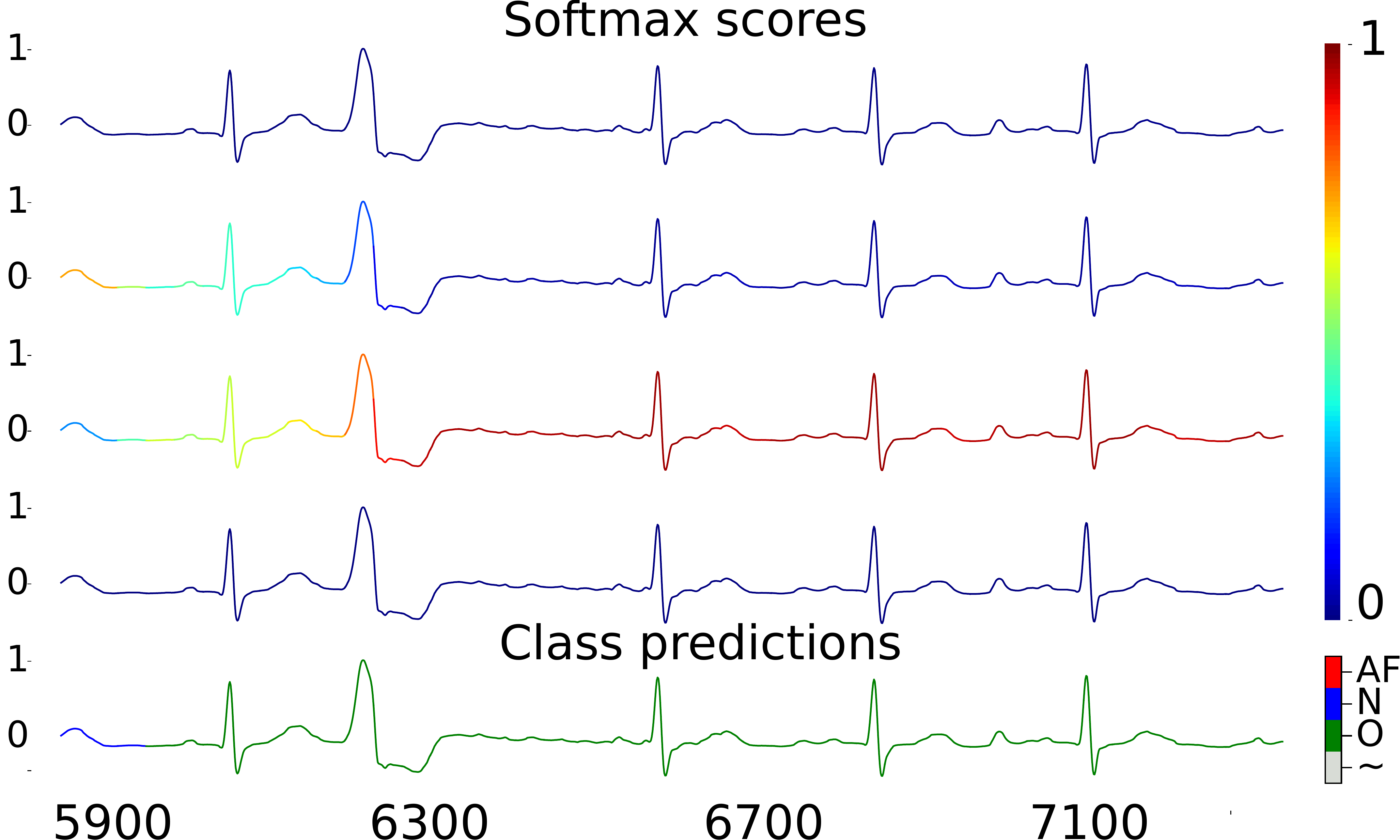}
	\caption{}
	\label{fig:lstm_timeplot2}
	\end{subfigure}
	\begin{subfigure}[b]{0.49\textwidth}
	\includegraphics[width=1.0\linewidth]{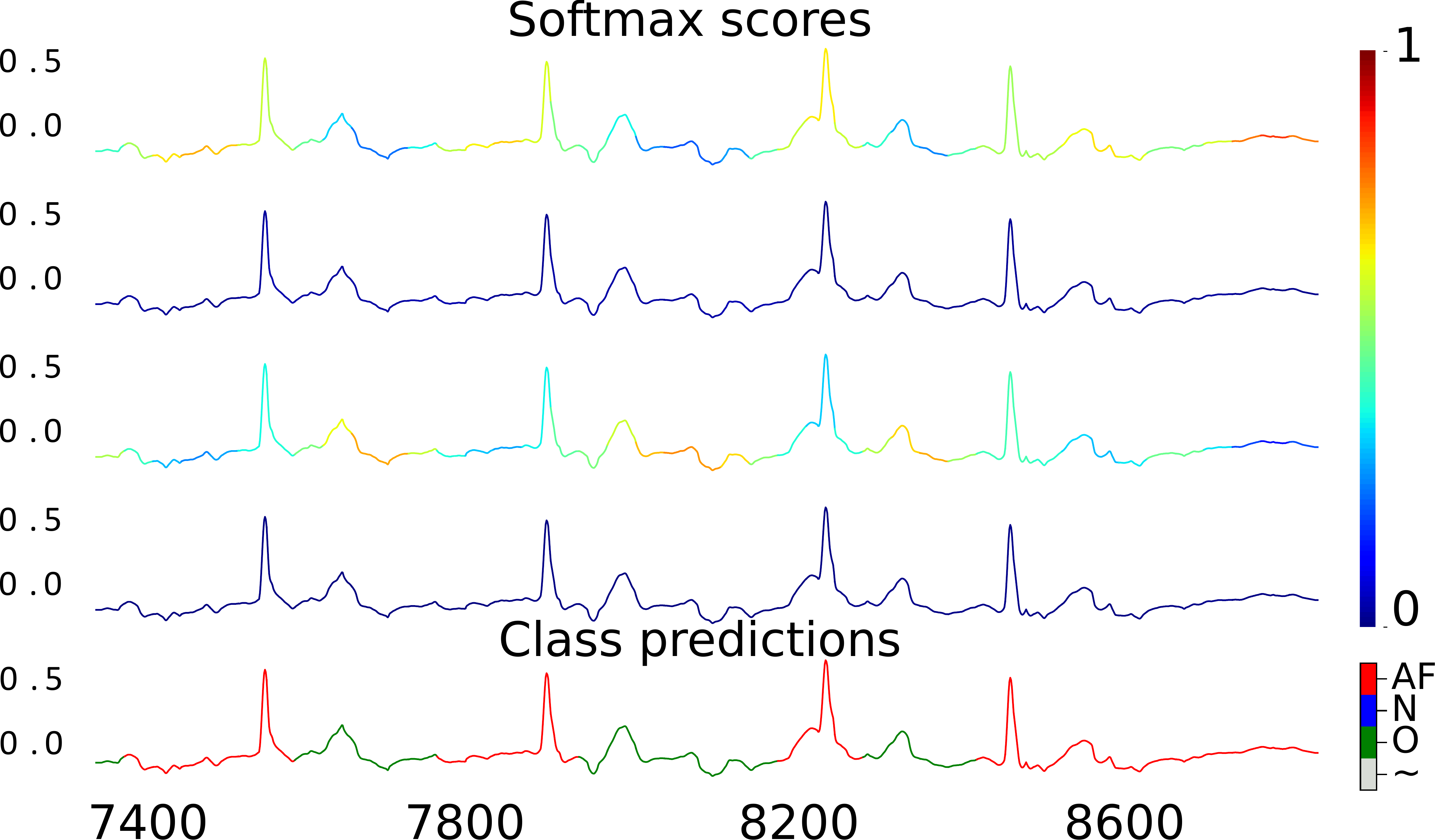}
	\caption{}
	\label{fig:lstm_timeplot3}
	\end{subfigure}
	\caption{5 second excerpt plot of intermediate class decisions for two examples of class Other being (a) correctly classified and (b) mistaken as class AF.}
	\label{fig:timeplot2}
\end{figure}

The following visualizations depict four `class decision over time' plots that were obtained for a ConvLSTM with 7 CNN layers and 16 hidden units. Since the output of the CNN was approximately 34 times smaller than the original input record, each time step represented a window of about 34 samples in the plot. The first example of Fig.~\ref{fig:lstm_timeplot0} shows a record excerpt that was correctly classified as Normal throughout the whole sequence. It can be observed that the prediction confidence slightly decreased with the beginning of a noisy event but that no class prediction switch was caused. The AF record of Fig.~\ref{fig:lstm_timeplot1}, on the contrary, was initially classified as class Other until a beat with missing P wave was encountered and the prediction switched to class AF. Generally, the plot of class Other rhythm examples allowed for the easiest interpretation. The 5 second excerpt of Fig.~\ref{fig:lstm_timeplot2}, for instance, clearly visualizes a strong enhancing of the class Other softmax score after the detection of an abnormal beat in the sequence. The last example of Fig.~\ref{fig:lstm_timeplot3} finally illustrates a failure case where a record with the class Other was confused as AF rhythm. Concerning this record, the network apparently permanently switched between the two class decisions Other and AF which likely originated from the fact that irregular RR intervals are a common feature for both classes. 


\begin{figure}
	\centering
	\begin{subfigure}[b]{0.49\textwidth}
	\includegraphics[width=1.0\linewidth]{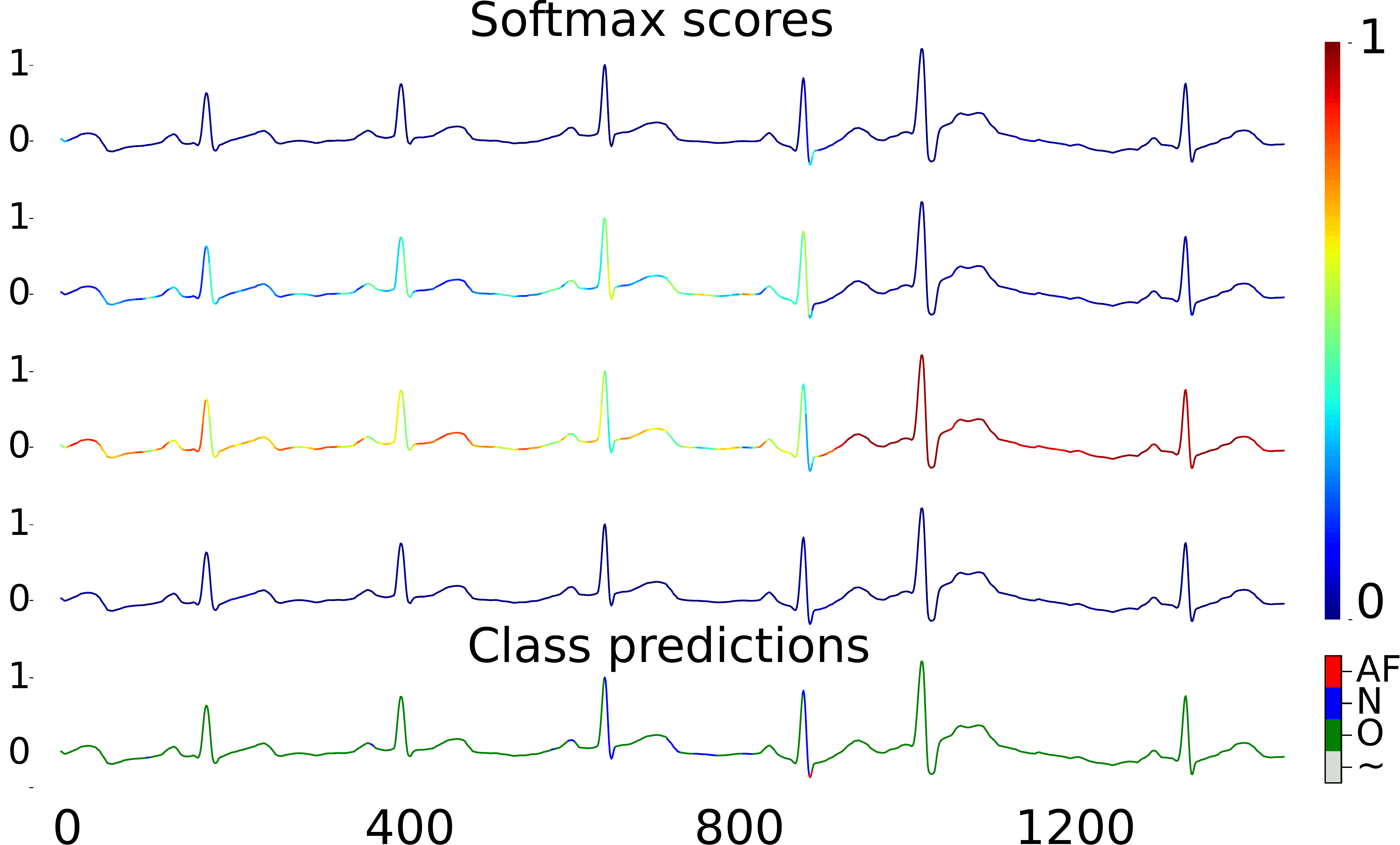}
	\caption{}
	\end{subfigure}
	\begin{subfigure}[b]{0.49\textwidth}
	\includegraphics[width=1.0\linewidth]{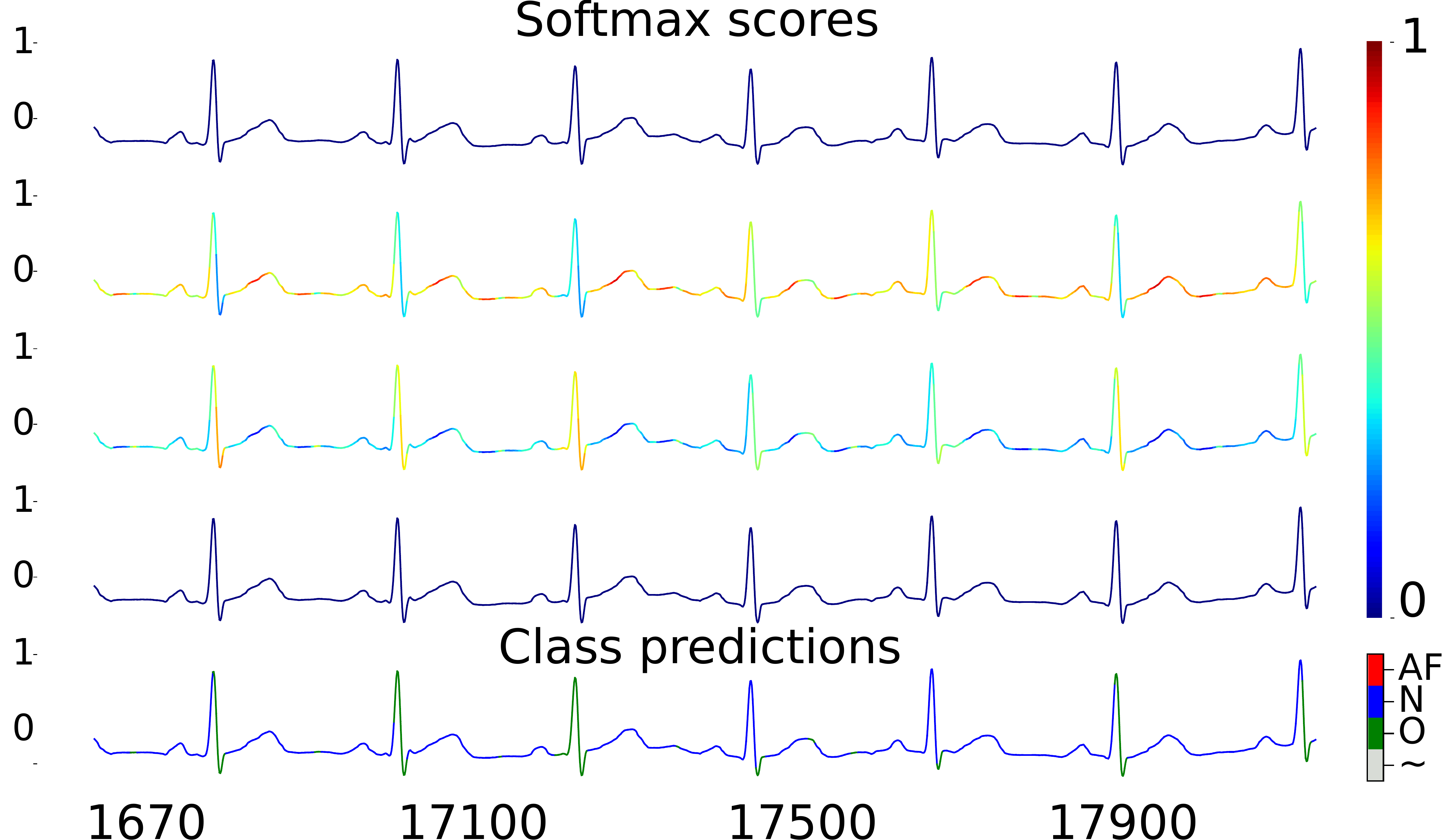}
	\caption{}
	\end{subfigure}
	\caption{Class decision plot for a shallower CNN module with only 4 layers showing the first 5 seconds (a) and the last 5 seconds (b) of an Other rhythm record that was incorrectly classified as Normal. It is likely that the misclassification was caused by limited memory capacities of the LSTM. Apparently, the class decision switched back to class normal at the end of the record, indicating that the cell `forgot' the detection of class Other patterns earlier in the sequence.} 
	\label{fig:lstm_timeplot4}
\end{figure}

When studying an even shallower ConvLSTM consisting of only 4 CNN and 2 LSTM layers with 128 hidden units, each time step represented a much smaller window of only 4 samples. Given that the LSTM processed a sequence of up to 4535 time steps, it is possible that the network had difficulties to remember salient patterns over long periods of time (e.g. if abnormal beats were encountered at the first part of the record). The example of Fig.~\ref{fig:lstm_timeplot4} depicts a class Other record where 
such a long-range memory issue likely have caused a misclassification as class Normal. Even though the class decision was performed correctly after the occurrence of abnormal morphologies, the decision switched back to class Normal at some point at the end of the sequence.

\subsection{Shift Perturbation Mask Visualizations} 
\begin{figure}
	\centering
	\begin{subfigure}[b]{1.0\textwidth}
	\includegraphics[width=1.0\linewidth]{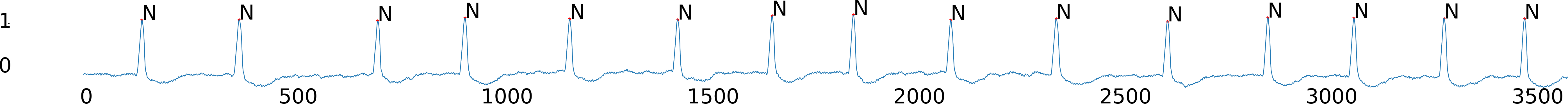}
	\label{fig:file209_start441775_end445375}
	\end{subfigure}
	\begin{subfigure}[b]{1.0\textwidth}
	\includegraphics[width=1.0\linewidth]{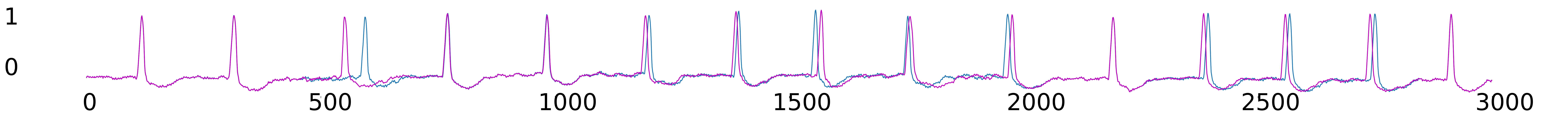}
	\label{fig:file209_target2_pred2_warped_deformation_epoch750}
	\end{subfigure}
	\begin{subfigure}[b]{1.0\textwidth}
	\includegraphics[width=1.0\linewidth]{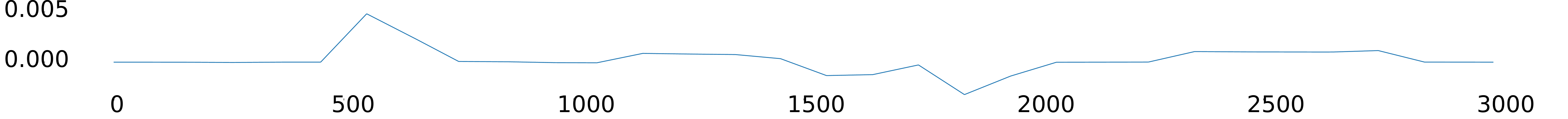}
	\label{fig:file209_target2_pred2_warped_perturbed_epoch799}
	\end{subfigure}
	\caption{MIT-BIH file 210 with meaningful perturbation for an AF example. Subfigures show the beat target annotation (top), the original signal in blue and the shifted signal in magenta (middle row), and finally the corresponding shift values of the upsampled pertubation grid (bottom). The perturbation yielded softmax scores of 0.00, 0.83, 0.17, and 0.00 for the classes AF, N, O, and Noisy using L1 and TV coefficients of 0.2 and 0.1.}
	\label{fig:mask1}
\end{figure}

\begin{figure}[H]
	\centering
	\begin{subfigure}[b]{1.0\textwidth}
	\includegraphics[width=1.0\linewidth]{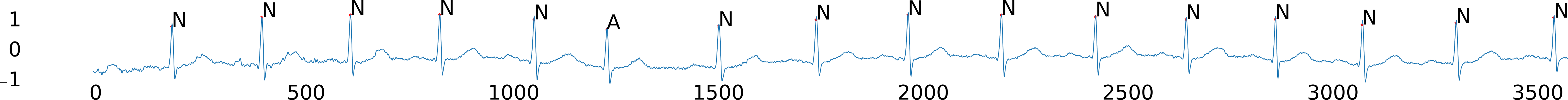}
	\label{fig:file209_start441775_end445375}
	\end{subfigure}
	\begin{subfigure}[b]{1.0\textwidth}
	\includegraphics[width=1.0\linewidth]{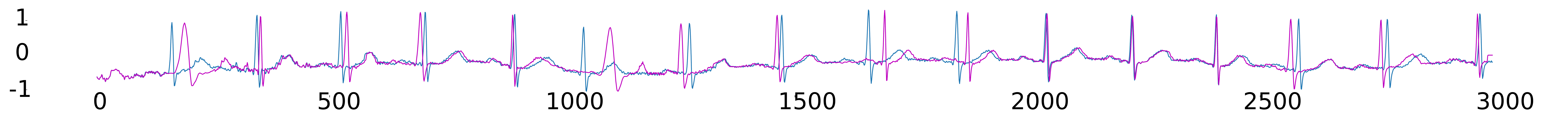}
	\label{fig:file209_target2_pred2_warped_perturbed_epoch799}
	\end{subfigure}
	\begin{subfigure}[b]{1.0\textwidth}
	\includegraphics[width=1.0\linewidth]{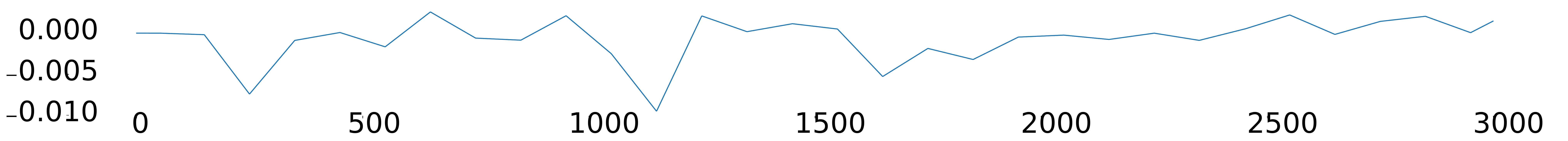}
	\label{fig:file209_target2_pred2_warped_deformation_epoch750}
	\end{subfigure}
	\caption{MIT-BIH file 209 with successful perturbation for a premature beat example of an class Other record. Resulting softmax scores were 0.01, 0.92, 0.03, and 0.04 for the classes AF, N, O, and Noisy using L1 and TV coefficients of 0.4 and 0.1.}
	\label{fig:mask2}
\end{figure}

\begin{figure}[H]
	\centering
	\begin{subfigure}[b]{1.0\textwidth}
	\includegraphics[width=1.0\linewidth]{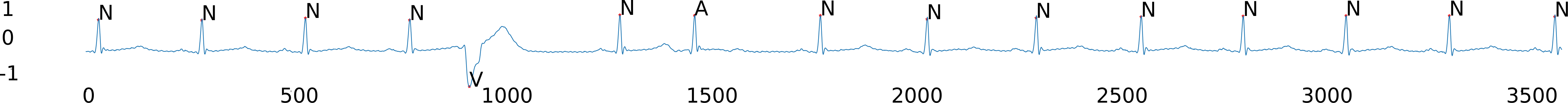}
	\label{}
	\end{subfigure}
	\begin{subfigure}[b]{1.0\textwidth}
	\includegraphics[width=1.0\linewidth]{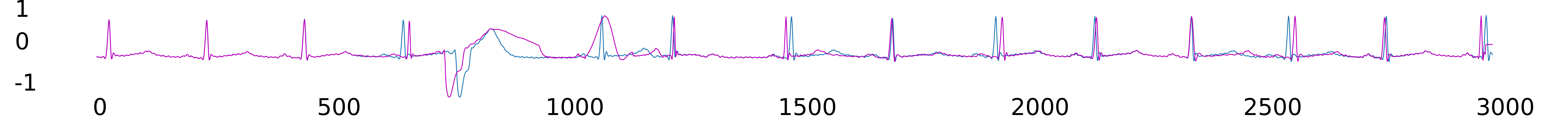}
	\label{}
	\end{subfigure}
	\begin{subfigure}[b]{1.0\textwidth}
	\includegraphics[width=1.0\linewidth]{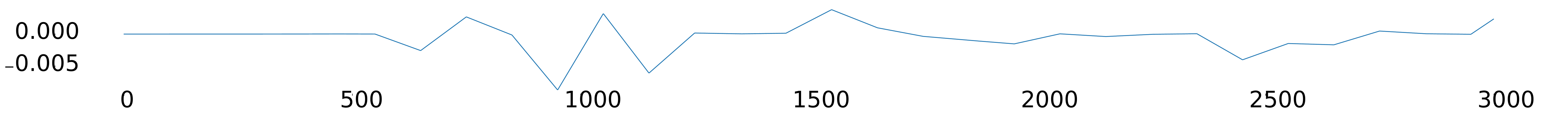}
	\label{}
	\end{subfigure}
	\caption{MIT-BIH file 205 with uninterpretable perturbation of an class Other record that surprisingly led to a class Normal prediction with softmax scores of 0.00, 0.98, 0.01, and 0.00 for the classes AF, N, O, and Noisy. The L1 and TV coefficients were set to 0.2 and 0.1.}
	\label{fig:mask4}
\end{figure}

As visible in Fig.~\ref{fig:mask1} and Fig.~\ref{fig:mask2}, the application of perturbation masks succeeded to identify irregular located beats for both AF and Other rhythms. The concept of `attention' in this context 
was expressed as the amount of shift that minimized the objective function of Eq.~\ref{eq:occlusionmask}. In other words, irregular beats of the input were supposed to be shifted in such an extent, that the network prediction switched from AF (or Other) to Normal (resulting from a maximal drop of the softmax score of e.g. AF when shifting the input samples with the given perturbation mask).
The records of Fig.~\ref{fig:mask1} and Fig.~\ref{fig:mask2} were finally successfully predicted as rhythm class Normal after the most prominent rhythm irregularities were removed by the perturbation mask. Especially the mask of the second example, however, also exhibits some higher shift values for beats of regular rhythm. Less interpretable is the perturbation of Fig.~\ref{fig:mask4} which was confidently predicted as class Normal even though the resulting perturbed signal has an unrealistic appearance.

\subsection{Hidden State and Gate Visualizations}
\label{sec:results:vis}
Since both the influence of LSTM parameters and the internal computations of LSTM cells remained hard to interpret, this sections aims at getting a better understanding of gate and state evolutions for a simple LSTM setup. The studied ConvLSTM (that yielded an F1 score of $0.83$) consisted of 15 convolutional layers and used only two LSTM units for the temporal aggregation. 
Given the many-to-one LSTM formulation, the classification layer of this setup received only two input activations which were the two entries of the last hidden state vector. Requiring the encoding of four classes by two values, the LSTM module apparently learned the following class specific hidden state representations: while the class AF prediction was associated with a very low activation for both neurons (color coded as dark blue), the class Normal showed high activations for both units (red), the class Other a high activation for the first and a low one for the second neuron, and lastly the class Noisy prediction was triggered by a combination of low and very low activations (dark and light blue). 

\begin{figure} 
\begin{minipage}{0.5\textwidth}
	\begin{subfigure}[b]{1\textwidth}
	\includegraphics[width=1.0\linewidth, trim={0 0.2cm 0 0}, clip]{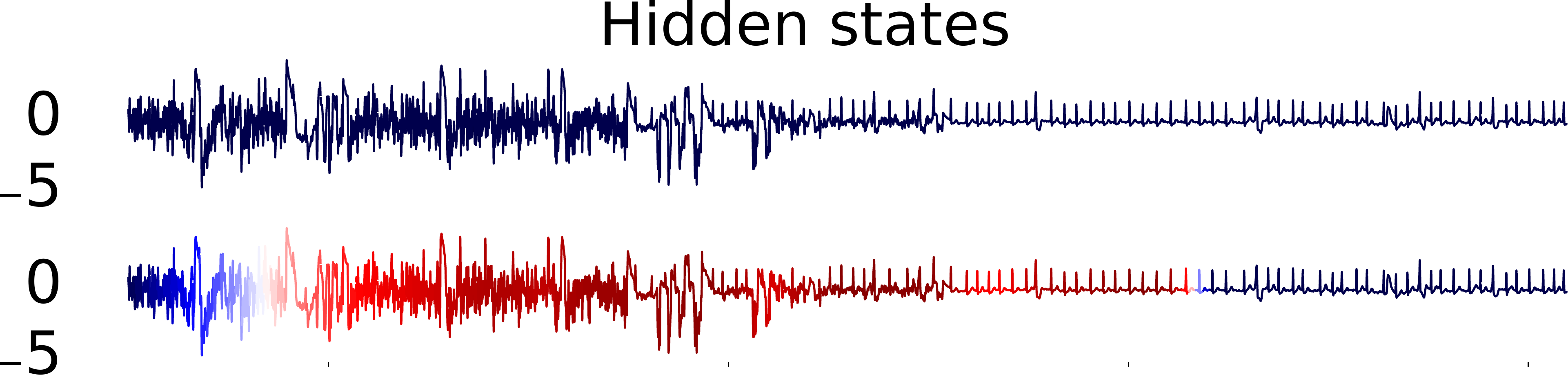}
	\caption{} 
	\label{fig:hidden_af}
	\end{subfigure}
	\begin{subfigure}[b]{1\textwidth}
	\includegraphics[width=1.0\linewidth, trim={0 0.9cm 0 0}, clip]{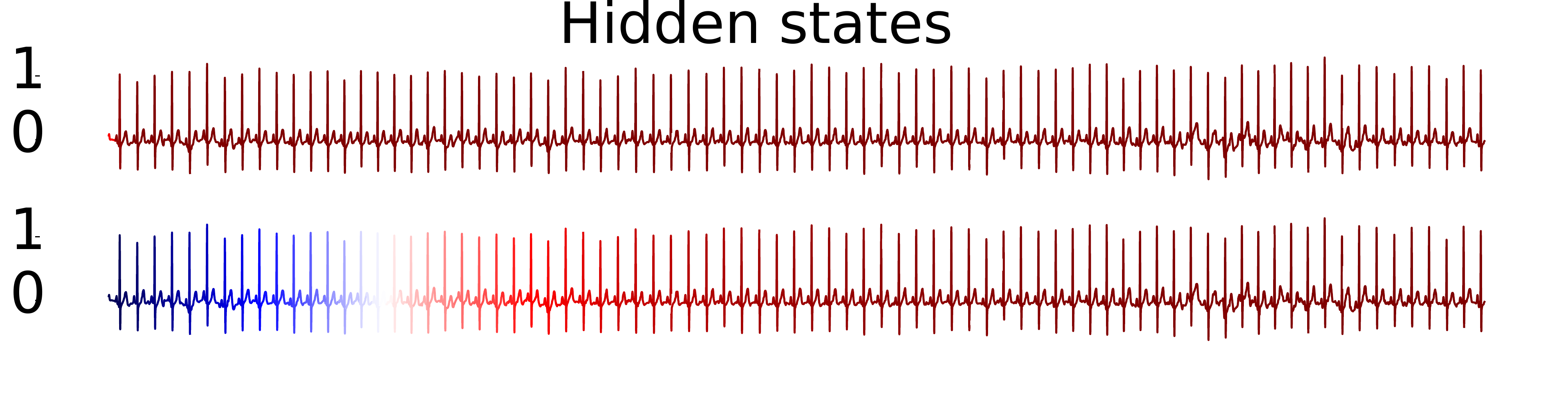}
	\caption{} 
	\label{fig:hidden_normal}
	\end{subfigure}
	\begin{subfigure}[b]{1\textwidth}
	\includegraphics[width=1.0\linewidth, trim={0 0.9cm 0 0}, clip]{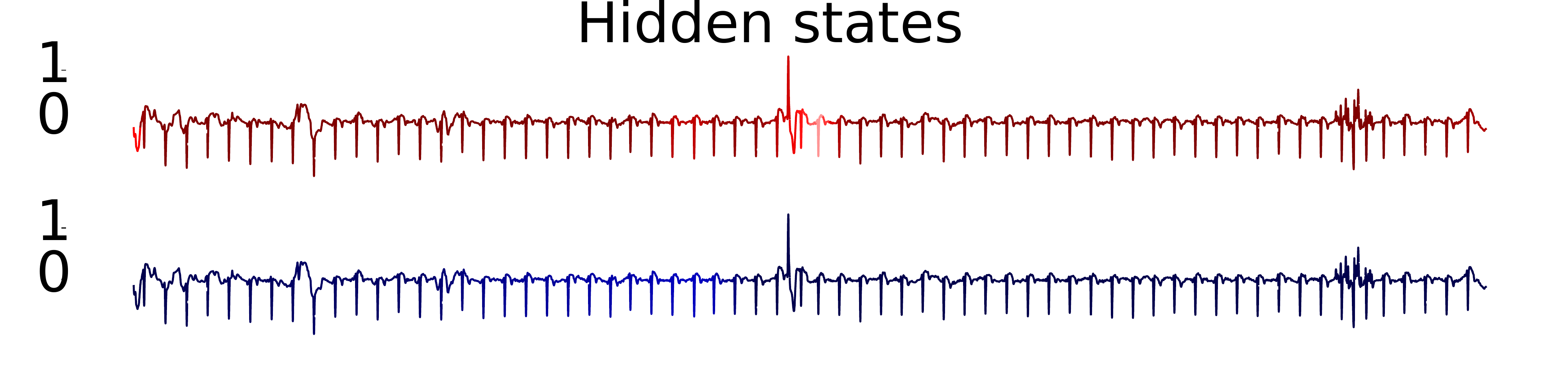}
	\caption{} 
	\end{subfigure}
	\begin{subfigure}[b]{1\textwidth}
	\includegraphics[width=1.0\linewidth, trim={0 0.3cm 0 0}, clip]{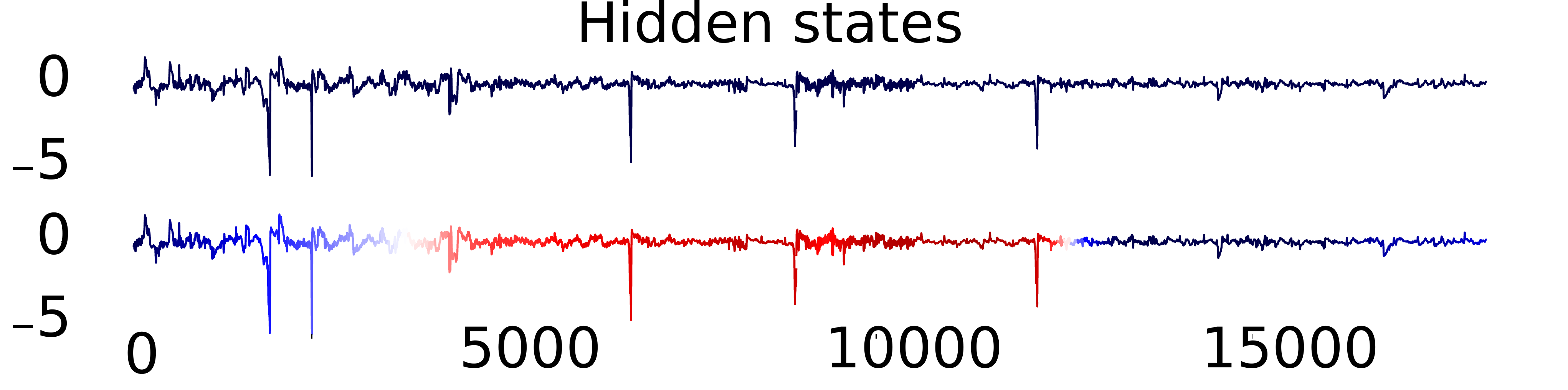}
	\caption{} 
	\label{fig:hidden_noisy}
	\end{subfigure}
 \end{minipage}
 \hfill
 \begin{minipage}{0.5\textwidth}
	\includegraphics[width=1.0\linewidth]{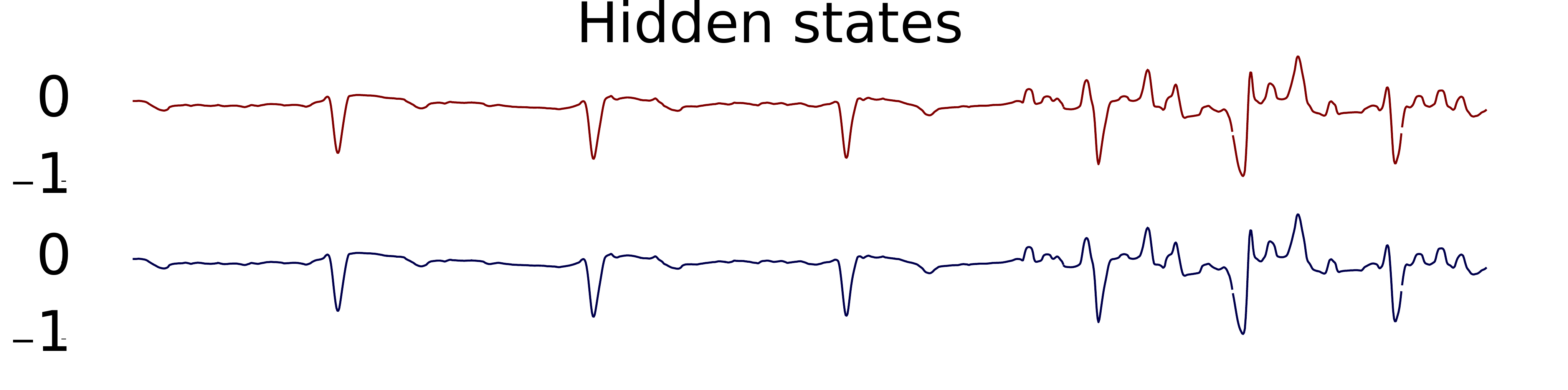}
	
	\begin{subfigure}[b]{1\textwidth}
	\includegraphics[width=1.0\linewidth]{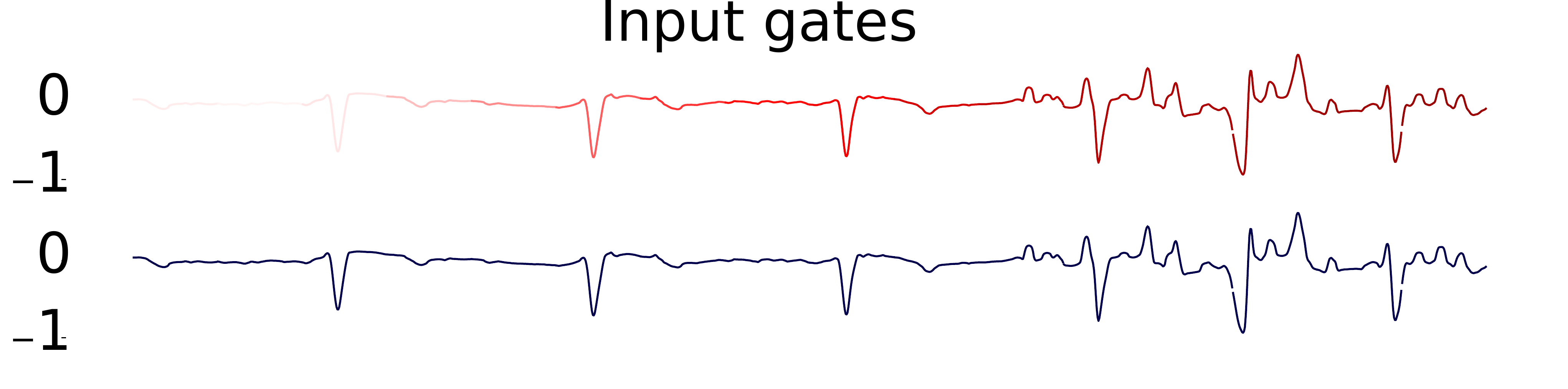}
	\end{subfigure}
	\begin{subfigure}[b]{1\textwidth}
	\includegraphics[width=1.0\linewidth]{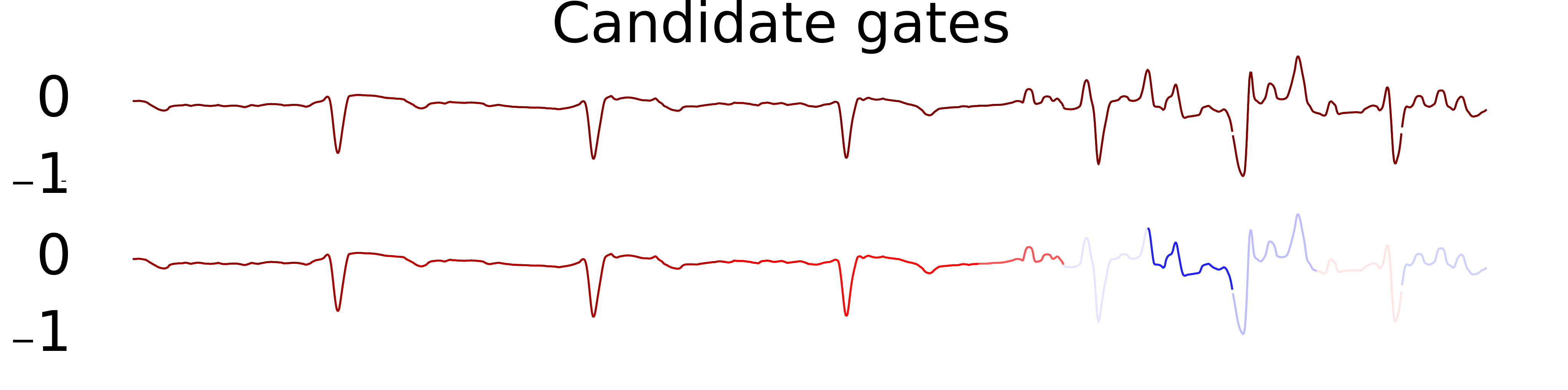}
	\end{subfigure}
	\begin{subfigure}[b]{1\textwidth}
	\includegraphics[width=1.0\linewidth]{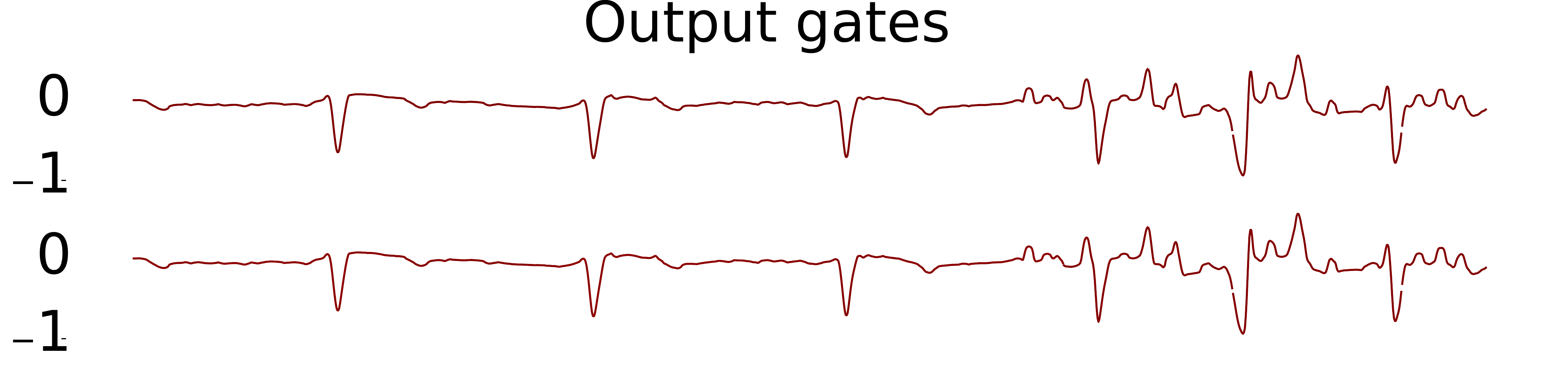}
	\end{subfigure}
	\begin{subfigure}[b]{1\textwidth}
	\includegraphics[width=1.0\linewidth]{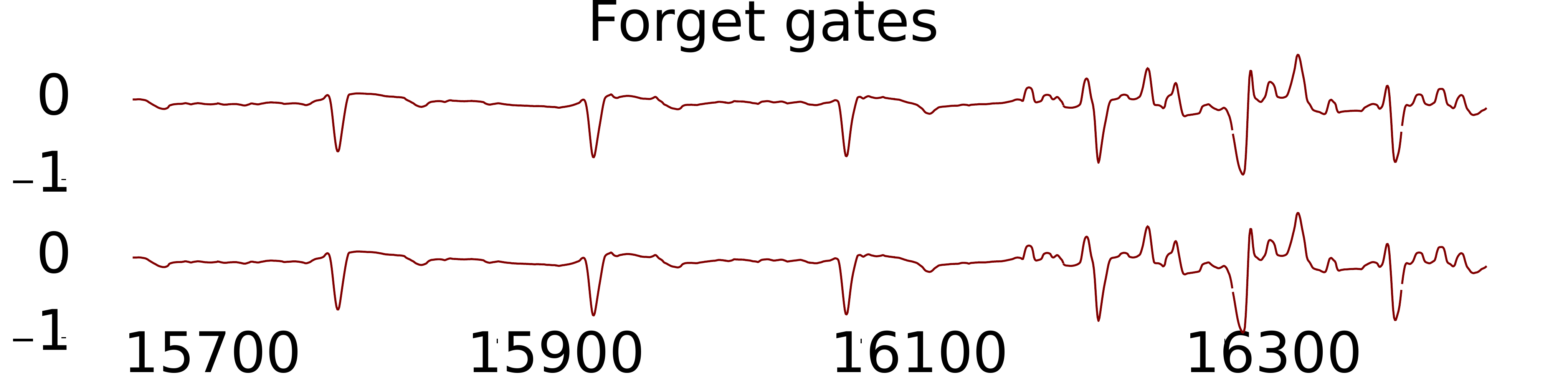}
	\end{subfigure}
 \end{minipage}
	\caption{Left: Visualization of LSTM hidden states with two hidden units for examples of classes AF (a), Normal (b), Other (c), and Noisy (d). Hidden state activations are color coded from high to low as dark red to dark blue. Examining the last hidden state for several examples, class specific encodings could be observed (showing e.g. two low activations for class normal records and a combination of high and low activations for class Other records). Right: Visualization of LSTM hidden states and gates with two hidden units for a 5 second excerpt of the preceding Other example. In this plot, a drop of candidate gate values can be observed after encountering a noisy record section.} 
	\label{fig:hidden_vis}
\end{figure}

Fig.~\ref{fig:hidden_vis} illustrates this encoding for four example cases. It can be observed that the `hidden state evolution over time plots' (see for instance Fig.~\ref{fig:hidden_af} and Fig.~\ref{fig:hidden_noisy}) are of similar appearance as the `class decision over time plots' studied earlier in this section. 
When sticking to the two unit class encoding hypothesis, the predictions for the record of class AF (depicted in Fig.~\ref{fig:hidden_af}) appear to have switched from initially AF, to Noisy, to Other, and finally back to the correct class prediction AF. 

Figure \ref{fig:hidden_vis} furthermore shows a zoomed 5 second excerpt of the depicted class Other record, which is displayed together with the corresponding input, candidate, output and forget gate values. In this example, the values of the forget gate and the output gate were consistently high, whereas the input and the candidate gates showed some changes over time.
While the drop of the candidate gate values with the beginning of a noisy record section appears reasonable, the remaining gate behaviors were more difficult to interpret. When recalling the hidden state and cell state definitions (which were $ c_{t} = f_{t}c_{t-1} + i_{t}g_{t}$ and $h_{t}=o_{t} \tanh (c_{t})$, respectively), the high values for both forget and output gates apparently kept the cell state entries from being overwritten by new inputs. In conclusion, even this simple LSTM module remained a black box concerning the internal gate and state modifications over time and no clear `responsibilities' of single neurons could be identified when analyzing the behavior of the cell for several examples.

\section{Performance Comparisons}
\begin{table}
\centering
\caption{8-fold cross validation performance comparison.}
\label{tab:evaluation}
\resizebox{\textwidth}{!}{%
\begin{threeparttable}[b]
\begin{tabular}{@{}rrrrrr@{}}
\toprule
Architecture & $F1_{AF}$ & $F1_{N}$  & $F1_{O}$  & $F1_{\sim}$  & $F1_{total}$  \\ \midrule
7 layer CNN+GAP & 0.79 ($ \pm $ 0.05) & 0.91 ($ \pm $ 0.01) & 0.75 ($ \pm $ 0.03) & 0.60 ($ \pm $ 0.07) &  0.82 ($ \pm $ 0.02) \\
7 layer CNN+GMP  & 0.78 ($ \pm $ 0.03) & 0.92 ($ \pm $ 0.01) & 0.76 ($ \pm $ 0.02) & 0.56 ($ \pm $ 0.08) &  0.82 ($ \pm $ 0.01) \\
7 layer CNN+1 layer LSTM, 16 hidden & 0.80 ($ \pm $ 0.03) & 0.92 ($ \pm $ 0.01)  &  0.78 ($ \pm $ 0.02)  & 0.65 ($ \pm $ 0.06) & 0.83 ($ \pm $ 0.02) \\
pretrained 7 layer CNN+2 layer LSTM, 16 hidden & 0.81 ($ \pm $ 0.02) & 0.92 ($ \pm $ 0.01) & 0.78 ($ \pm $ 0.02)& 0.65 ($ \pm $ 0.06) & 0.84 ($ \pm $ 0.02)\\
15 layer CNN+GMP  & 0.82 ($ \pm $ 0.02) & 0.92 ($ \pm $ 0.01) & 0.79 ($ \pm $ 0.02) & 0.63 ($ \pm $ 0.07) &  0.84 ($ \pm $ 0.01) \\
pretrained 15 layer CNN+1 layer LSTM, 4 hidden & 0.82 ($ \pm $ 0.03) & 0.92 ($ \pm $ 0.01) & 0.78 ($ \pm $ 0.02) & 0.62 ($ \pm $ 0.06) &  0.84 ($ \pm $ 0.02) \\ 
pretrained 15 layer CNN+1 layer LSTM, 64 hidden & 0.83 ($ \pm $ 0.03) & 0.92 ($ \pm $ 0.01) & 0.79 ($ \pm $ 0.02) & 0.64 ($ \pm $ 0.06) &  \textbf{0.85} ($ \pm $ 0.02) \\ 
pretrained 15 layer CNN+2 layer LSTM, 64 hidden  & 0.82 ($ \pm $ 0.03) & 0.92 ($ \pm $ 0.01) & 0.79 ($ \pm $ 0.02) & 0.64 ($ \pm $ 0.09) &  0.84 ($ \pm $ 0.02) \\
17 layer CNN+GMP & 0.81 ($ \pm $ 0.04) & 0.92 ($ \pm $ 0.01) & 0.78 ($ \pm $ 0.02) & 0.65 ($ \pm $ 0.04) &  0.83 ($ \pm $ 0.02) \\
17 layer CNN+AttentionGates, mean vote & 0.81 ($ \pm $ 0.04) & 0.92 ($ \pm $ 0.01) & 0.78 ($ \pm $ 0.02) & 0.63 ($ \pm $ 0.09) &  0.84 ($ \pm $ 0.02)  \\
\bottomrule
\end{tabular}
\end{threeparttable}
}
\end{table}

Table \ref{tab:evaluation} summarizes the presented 8-fold cross validation scores for a selection of experiments. The best average F1 score of $0.85$ was obtained by a single layer LSTM consisting of 64 hidden units that was stacked on top of a pretrained CNN with 15 layers. Especially for deep CNN setups, a pretraining of CNN parameters showed to be beneficial. While the hidden state capacity appeared to have a large influence on the performance for the 4 layer CNN setup (obtaining a score of $0.69$ when using 16 hidden units and $0.75$ when increasing the number of hidden units to 64), a smaller influence was observed for ConvLSTMs with deeper CNN modules. 
Moreover, it was found that simple global pooling CNN architectures performed surprisingly well. 
For both 7 layer and 15 layer CNN modules the performances were only slightly worse than those of ConvLSTM networks using the same amount of CNN layers. Finally, the application of attention gates in a 17 layer deep CNN network did only insignificantly improve the baseline CNN performances from $0.83$ to $0.84$.

The confusion matrices of Table \ref{tab:confusion1} and Table \ref{tab:confusion2} (which are the summed confusion matrices over all 8 folds) compare the class prediction distributions of global max pooling and LSTM setups for a 7 layer and a 15 layer CNN module, respectively. The depicted distribution suggests that the \textbf{7 layer CNN+GAP} setup was the most sensitive to AF cases and at the same time the least sensitive for class Other rhythms. An opposing behavior was observed for the \textbf{15 layer CNN+LSTM} setup which detected the fewest AF and the most Other rhythm records. Moreover, it becomes apparent that for both global pooling and LSTM variants the number of correctly classified Other records increased significantly with growing network depth.

An examination of the overall F1 scores for the class AF (which were 0.79 for \textbf{7 layer CNN+GAP}, 0.81 for \textbf{7 layer CNN+LSTM}, 0.82 for \textbf{15 layer CNN+GMP}, and 0.83 for \textbf{15 layer CNN+LSTM}), shows that the high sensitivity of the shallow global pooling setup came along with a comparably high number of false positive predictions. Consequently, LSTMs yielded superior scores concerning the F1 performance metric. A study of the overall F1 scores for the class Other (which were 0.75 for \textbf{7 layer CNN+GAP}, 0.78 for \textbf{7 layer CNN+LSTM}, 0.79 for \textbf{15 layer CNN+GAP}, and 0.79 for \textbf{15 layer CNN+LSTM}) again suggested that the accuracy was more impacted by the network depth than by the choice of temporal aggregation strategy.

{\small
\begin{table} 
\caption{Confusion matrices for \textbf{7 layer CNN+GMP} (left) and \textbf{pretrained 7 layer CNN+2 layer LSTM, 16 hidden} (right).}
\label{tab:confusion1}
\resizebox{0.9\textwidth}{!}{%
\begin{minipage}[b]{0.5\linewidth}\centering
\begin{tabular}{l|l|c c c c|c}
\multicolumn{2}{c}{}&\multicolumn{4}{c}{Ground truth}&\\
\cline{3-6}
\multicolumn{2}{c|}{}&AF&N&O&$\sim$&\multicolumn{1}{c}{Total}\\
\cline{2-6}
\multirow{4}{*}{\rotatebox{90}{Prediction}}&  AF & 688 & 70 & 236 & 21 & 1015\\
& N & 13 &  4707 & 423  & 57 & 5200\\
&  O & 48 & 225 & 1663 & 21 & 1957\\
&$\sim$ & 9 & 74  & 93 & 180 & 356\\
\cline{2-6}
\multicolumn{1}{c}{} & \multicolumn{1}{c}{Total} & \multicolumn{1}{c}{758} & \multicolumn{    1}{c}{5076} & \multicolumn{1}{c}{2415} & \multicolumn{1}{c}{279} & \multicolumn{1}{c}{8528}\\
\end{tabular}
\end{minipage}
\hspace{0.5cm}
\begin{minipage}[b]{0.5\linewidth}
\centering
\begin{tabular}{l|l|c c c c|c}
\multicolumn{2}{c}{}&\multicolumn{4}{c}{Ground truth}&\\
\cline{3-6}
\multicolumn{2}{c|}{}&AF&N&O&$\sim$&\multicolumn{1}{c}{Total}\\
\cline{2-6}
\multirow{4}{*}{\rotatebox{90}{Prediction}}&   AF & 637 & 28 & 134 & 14 & 813\\
&N & 20 & 4795 & 487 & 50 & 5156\\
& O & 83 & 204 & 1740 & 22 & 2287\\
& $\sim$ & 18 & 49 & 54 & 193 & 272\\
\cline{2-6}
\multicolumn{1}{c}{} & \multicolumn{1}{c}{Total} & \multicolumn{1}{c}{758} & \multicolumn{    1}{c}{5076} & \multicolumn{1}{c}{2415} & \multicolumn{1}{c}{279} & \multicolumn{1}{c}{8528}\\
\end{tabular}
\end{minipage}
}
\end{table}
}

\begin{table} 
\caption{Confusion matrices for \textbf{15 layer CNN+GMP} (left) and \textbf{pretrained 15 layer CNN+1 layer LSTM, 64 hidden} (right).}
\label{tab:confusion2}
\resizebox{0.9\textwidth}{!}{%
\begin{minipage}[b]{0.5\linewidth}\centering
\begin{tabular}{l|l|c c c c|c}
\multicolumn{2}{c}{}&\multicolumn{4}{c}{Ground truth}&\\
\cline{3-6}
\multicolumn{2}{c|}{}&AF&N&O&$\sim$&\multicolumn{1}{c}{Total}\\
\cline{2-6}
\multirow{4}{*}{\rotatebox{90}{Prediction}}&  AF & 641 & 37 & 124 & 11 & 813\\
&N & 14 &  4703 & 379  & 60 & 5352\\
&O & 90 & 301 & 1864 & 32 & 2049\\
&$\sim$ & 13 & 35 & 48 & 176 & 314\\
\cline{2-6}
\multicolumn{1}{c}{} & \multicolumn{1}{c}{Total} & \multicolumn{1}{c}{758} & \multicolumn{    1}{c}{5076} & \multicolumn{1}{c}{2415} & \multicolumn{1}{c}{279} & \multicolumn{1}{c}{8528}\\
\end{tabular}

\end{minipage}
\hspace{0.5cm}
\begin{minipage}[b]{0.5\linewidth}
\centering
\begin{tabular}{l|l|c c c c|c}
\multicolumn{2}{c}{}&\multicolumn{4}{c}{Ground truth}&\\
\cline{3-6}
\multicolumn{2}{c|}{}&AF&N&O&$\sim$&\multicolumn{1}{c}{Total}\\
\cline{2-6}
\multirow{4}{*}{\rotatebox{90}{Prediction}}& AF  & 633 & 21 & 99 & 13 & 766\\
& N & 18 & 4706 & 403& 59 & 5186\\
& O & 96 & 318 & 1869 & 35 & 2318\\
& $\sim$ & 11 & 31 & 44 & 172 & 258\\
\cline{2-6}
\multicolumn{1}{c}{} & \multicolumn{1}{c}{Total} & \multicolumn{1}{c}{758} & \multicolumn{    1}{c}{5076} & \multicolumn{1}{c}{2415} & \multicolumn{1}{c}{279} & \multicolumn{1}{c}{8528}\\
\end{tabular}
\end{minipage}
}
\end{table}

\chapter{Discussion}
\label{sec:discussion}

\section{Are ConvLSTMs the Winners of the CinC Challenge?}
It does not directly follow that the excellent performance of our ConvLSTM network of $F1=0.85$ on the CinC training set would be the best ranking performance on the official test set (for which a best score of $F1=0.83$ was reported). A drop of performance was, for instance, noted by Warrick et al.~\cite{warrick2017} who achieved a score of $F1=0.83$ for a 10-fold cross validation and a score of only $0.80$ on the test set. Consequently, a fair comparison to other state-of-the-art approaches, so far, is only limited possible and a submission of our best performing model intended. 

Assuming a similar performance drop of about $0.03$, the proposed ConvNet would, however, yield a satisfying performance while keeping the classification pipeline much simpler than most of the participating teams (see Sec.~\ref{sec:recentwork}). As intended, the proposed setup not only avoided additional post-processing steps but also the requirement of complex feature engineering pipelines. Furthermore, it is to be assumed that both an ensemble of multiple, independently trained ConvLSTMs and the application of data augmentation could further improve the performances. 

\paragraph{Are pure CNNs better classifiers than ConvLSTMs?}

The experiments of this work suggest that the complex temporal aggregation of LSTMs (which adds many learnable parameters) did not significantly outperform global pooling strategies (which, on the contrary, basically discard information). Even if this observation is surprising, it is in line with the results reported by Zihlmann et al.~\cite{zihlmann2017} who discovered that for a 24 layer CNN setup, LSTMs outperformed GAP only in case data augmentation was employed. 
Still, a more obvious superiority of LSTMs was observed for the 4 and 7 layer CNN modules, where LSTMs apparently provided a `smarter' temporal aggregation of local beat features than the competing global pooling layers. However, with a growing receptive field (that comes along with deeper networks), CNNs seemed to successfully take over the task of capturing temporal long-range dependencies (like rhythm change informations). This interpretation was emphasized by class activation map visualizations which showed that CNN networks successfully managed to detect irregular rhythm sections. The choice of the aggregation strategy certainly also depends on the definition of the ground truth annotations. If, for instance, records where classified as AF as soon as some f oscillations or irregular RR interval changes were included (features that could be well detected by CNN layers) or whether additional knowledge of e.g. rhythm change durations needed to be captured (which potentially could be better performed by LSTM layers). 

Furthermore, the stacking of multiple LSTM layers is supposed to yield higher level temporal features at different time scales \cite{malhotra2015},\cite{karpathy2015}. For the application of character-level language modeling, Karpathy \cite{karpathy2016} published some rules of thumbs telling that usually 2 to 3 layers perform well and the number of hidden units should be chosen according to the amount of available data. Still, throughout the experiments of this work, neither the number of LSTM layers nor the amount of hidden units had a large impact on the rhythm classification performance. 

Another study was recently published by Yin et al.~\cite{yin2017}, who compared CNN and RNN performances for the application of natural language processing. They concluded that no answer can be found to the question which setup generally performs best and advised to use CNNs for classification tasks (like sentiment analysis, where a class is usually determined by some key features) and RNNs for sequence modeling applications (like language modeling). Nevertheless, Yin et al. also referenced related works where RNNs performed well for document-level sentiment classification \cite{tang2015} or gated CNNs outperformed LSTMs on language modeling tasks \cite{dauphin2016}. 

\paragraph{Benefits of bidirectional LSTMs and GRUs?}
It was assumed that bidirectional LSTMs could improve the classification accuracy particularly for records where pathological findings were observed in the first part of the time sequence. Given that backward LSTMs in such cases need to remember detected events over less time steps, the incorporation of the backward hidden state was expected to provide beneficial supplementary information. The experiments of this work, however, did not confirm this hypothesis (neither the concatenation of hidden states for both directions nor the extraction of hidden states at the central time step).

Arkhipenko et al.~\cite{arkhipenko2016} recently evaluated GRUs for the task of sentiment analysis and found GRUs to outperform both LSTMs and pure CNNs. As introduced in Sec.~\ref{sec:methods:lstm}, the GRU cell uses a merged formulation of the hidden state and the cell state and also reduces the number of gates from four to three. However, since performances in first experiments did not seem to improve and as other empirical studies also came to the conclusion that LSTMs and GRUs often perform comparably well (see e.g. Chung et al. \cite{chung2014}), 
we did not follow up on further GRU investigations.

\paragraph{Rise and fall of LSTMs?}
Some researchers even state ``drop your RNN and LSTM, they are no good'' \cite{culurciello2018}, claiming that companies like Google and Facebook would start to replace RNN architectures with attention based models. Culurciello \cite{culurciello2018} argues that the use of LSTMs should be generally avoided since recurrent architectures are not only limited in memory capacities but also computationally expensive. Hierarchical attention models, on the contrary, could integrate more time steps and would require shorter paths in the backpropagation pass (where the length in tree hierarchies is proportional to the logarithm of the tree depth 
while standard RNNs propagate the error through all time steps of the sequence) \cite{culurciello2018}.

\section{Can Attention Visualizations Support AF Diagnosis?}
In this work, we aimed at developing visualization tools to better understand the internal processes of neural network models and to support clinicians focusing on meaningful ECG sections during AF diagnosis. In the following, it will be discussed which attention maps were the most promising and which challenges might remain.

\paragraph{Class activation maps}
The idea of class activation maps proved to be promising and stunningly simple. The attention maps of the previous section gave visual proof that global pooling CNNs can facilitate a precise and at the same time interpretable detection of pathological events in long term ECGs. Still, those CAMs only provide a rough approximation to the regions of the highest importance for the CNN classification. As it has been shown, when several pathological episodes occur in a single window, one or more of these episodes could be missed with the CNN focusing on another particular episode. 

Approaches to improve localization abilities of class activation maps were extensively discussed for weakly supervised object localization applications. An example work that managed to better capture the full extent of detected objects was 2017 published by Dahun et al.~\cite{dahun2017}. In their work, the authors proposed to suppress relevant CNN neurons of highest activations in a second training phase in order to encourage a network to look for further class evidences. They found that this `two phase learning' resulted in more accurate heat maps of localized objects. Concerning the ECG classification task of this thesis, such a two phase learning could potentially allow for the detection of more pathological episodes. 

\paragraph{LSTM class decision plots}
The plot of intermediate class decisions provided attention-like maps for recurrent neural networks. Beside highlighting salient record sections, it allowed for a better understanding of the sequential input processing and also helped to identify long-range memory difficulties (which were often indicated by alternating class decisions at the end of the record plot). 

\paragraph{Shift perturbations}
The `attention maps' that resulted from the computation of perturbation masks allowed for the examination of saliency without requiring any modifications (not to mention any internal parameter extractions) of the underlying model. By replacing the original `occlusion mask' formulation (where samples were occluded by constant values, noise, or blur) by shift perturbations, we presented a novel 
approach for the detection of pathological episodes in AF and class Other records. The concept of perturbation masks proved to be an interesting tool for the manipulation of network decisions and might be beneficial for realistic data augmentation. Nevertheless, it is unlikely that perturbation mask attention visualizations can actually support clinical diagnosis since the optimization process appeared unstable and the applicability was limited to a selection of rhythm types.

\section{Are Attention Mechanisms Beneficial for the Training Process?}

Inspired by the human visual attention system, attention networks learn to focus on important input details and to fade-out irrelevant background information. The attention gated CNN architecture studied in this work was therefore expected to first identify important rhythm features on a global scale and to afterward focus on local morphology information of salient beats (to e.g. more successfully discriminate between AF and Other rhythm beats which might both show sections of irregular RR intervals at the coarsest scale). 

\paragraph{Attention gating in CNN models}
In fact, such a focusing on particular beats could be observed for the gated attention map examples of Fig.~\ref{fig:meanvote_conv7_conv9} (at least for the cases of AF and Normal rhythm). Contrary to the attention maps reported by Schlemper et al.~\cite{schlemper2018}, highlighted beats were, however, not located in the same rhythm section that had been before detected by the global classification paths. 

The reason for this observation can be probably found in the additive attention weight definition of Eq.~\ref{eq:attentiongate}, where compatibility scores are computed between the intermediate and the last layer's feature map. Since ECG records show repetitive patterns throughout the whole record, it is possible that feature vectors of high compatibility scores less likely corresponded to beats included in the detected region of the global feature map (whereas Schlemper et al.~\cite{schlemper2018} processed images where objects could be more easily discriminated from background regions). Moreover, attention maps that were extracted for class Other records hardly showed any high activations, which implies that fine-scale features for those cases did not contribute to the prediction.

Summarizing, it was found that the incorporation of attention gates could slightly improve the F1 score of the baseline CNN from $0.83$ to $0.84$ but that resulting attention maps where less intuitive than simple class activation maps of shallower networks (with similar performances).
\chapter{Conclusion}
\label{sec:conclusion}

In this work, convolutional long short-term memory were proposed for the detection of AF rhythms in single-lead ECG recordings. Combining the benefits of both CNN and LSTM architectures, the network successfully captured features of morphology and rhythm changes from raw ECG records while not requiring any pre- or post-processing. Yielding an F1 score of $0.85$ for an 8-fold cross validation on the CinC 2017 challenge training data, the network performed similarly to the top ranked challenge approaches with a score of $0.83$ on the unavailable test set. To allow for a performance assessment and a fair comparison with other state-of-the-art approaches, a submission of our best model to the PhysioNet community is intended. Comparing the temporal feature aggregation abilities of LSTMs and global pooling layers, a slight superiority of LSTMs was observed for shallow CNN setups. However, when increasing the depth of CNN architectures for the feature extraction, no significant performance differences could be reported.

In addition, various attention visualization techniques were presented for CNN as well as LSTM architectures. By successfully highlighting pathological episodes of morphology or rhythm irregularities, attention maps proved to have a great potential to support clinicians for cardiac diagnosis in long-term ECGs. It is assumed that attention mechanisms can help to speed up diagnosis, to yield better classification transparency and to reassess cases of low prediction certainty. The extension of a standard CNN network with additional trainable attention parameters only insignificantly improved the performance from $F1=0.83$ to $F1=0.84$. Resulting attention maps were more difficult to interpret than simple class activation maps and therefore appeared less helpful for the task of clinical diagnosis support.

In future research, we plan to adopt our model for the processing of ECG data that was acquired during magnetic resonance imaging (MRI). Given that the presence of a static magnetic field distorts the recordings of heart activity \cite{oster2015}, a special handling of noise and artifacts will be required to still enable arrhythmia detection. For this purpose, denoising strategies will be exploited using, for instance, autoencoder networks \cite{hinton2006} for an unsupervised pretraining. Moreover, it is suggested that a pretraining on the large PhysioNet database with subsequent transfer learning can improve the generalization abilities of models for the comparably small database.


\bibliographystyle{gerplain}
\footnotesize\bibliography{./includes/references}

\end{document}